%% file: main_neurips_camera.tex
\documentclass{article}
\newif\ifchecklist \checklistfalse  % NeurIPS camera-ready: \checklisttrue, arXiv: \checklistfalse

\providecommand{\cspstd}[1]{{\scriptsize$\pm#1$}}
\PassOptionsToPackage{numbers,compress}{natbib}

\usepackage[main, final]{neurips_2026}

\usepackage[utf8]{inputenc}
\usepackage[T1]{fontenc}
\usepackage{microtype}
\usepackage{graphicx}
\usepackage{subcaption}
\usepackage{booktabs}
\usepackage{hyperref}
\usepackage{url}
\usepackage{adjustbox}
\usepackage{xcolor}
\usepackage{colortbl}
\usepackage{pifont}
\usepackage{multirow}
\usepackage{amsmath}
\usepackage{amssymb}
\usepackage{amsfonts}
\usepackage{mathtools}
\usepackage{amsthm}
\usepackage{cancel}
\usepackage{nicefrac}
\usepackage[capitalize,noabbrev]{cleveref}
\usepackage[font=small,labelfont=bf]{caption}
\usepackage{algorithm}
\usepackage[noend]{algpseudocode}
\usepackage{makecell}

\theoremstyle{plain}

\theoremstyle{definition}

\theoremstyle{remark}

\usepackage{wrapfig}

\DeclareMathOperator*{\argmax}{arg\,max}
\hypersetup{
    colorlinks=true,
    citecolor=blue,
    linkcolor=black,
    urlcolor=teal
}

\title{Generative Recursive Reasoning}

\author{%
  \makebox[\textwidth][c]{%
    \textbf{Junyeob Baek}$^{1}$\thanks{Equal contribution} \quad 
    \textbf{Mingyu Jo}$^{1*}$ \quad 
    \textbf{Minsu Kim}$^{1,2}$
  } \\
  \vspace{2mm} \\
  \makebox[\textwidth][c]{%
    \textbf{Mengye Ren}$^{3}$ \quad 
    \textbf{Yoshua Bengio}$^{2,4}$ \quad 
    \textbf{Sungjin Ahn}$^{1,3}$\thanks{Correspondence to: Sungjin Ahn (\texttt{sungjin.ahn@kaist.ac.kr})}
  } \\
  \vspace{4mm} \\
  \makebox[\textwidth][c]{$^1$KAIST \quad $^2$Mila -- Qu{\'e}bec AI Institute} \\
  \makebox[\textwidth][c]{$^3$New York University \quad $^4$Universit{\'e} de Montr{\'e}al}
}

\begin{document}

\maketitle

\begin{abstract}

How should future neural reasoning systems implement extended computation? Recursive Reasoning Models (RRMs) offer a promising alternative to autoregressive sequence extension by performing iterative latent-state refinement with shared transition functions. Yet existing RRMs are largely deterministic, following a single latent trajectory and converging to a single prediction. We introduce \emph{Generative Recursive reAsoning Models (GRAM)}, a framework that turns recursive latent reasoning into probabilistic multi-trajectory computation. GRAM models reasoning as a stochastic latent trajectory, enabling multiple hypotheses, alternative solution strategies, and inference-time scaling through both recursive depth and parallel trajectory sampling. This yields a latent-variable generative model supporting conditional reasoning via $p_\theta(y \mid x)$ and, with fixed or absent inputs, unconditional generation via $p_\theta(x)$. Trained with amortized variational inference, GRAM improves over deterministic recurrent and recursive baselines on structured reasoning and multi-solution constraint satisfaction tasks, while demonstrating an unconditional generation capability. \href{https://ahn-ml.github.io/gram-website/}{https://ahn-ml.github.io/gram-website}

\end{abstract}

\section{Introduction}

A central question for future neural reasoning systems is how extended computation should be implemented.~Large autoregressive models typically scale reasoning by extending a sequence-generation process, whether intermediate computation is expressed explicitly as chain-of-thought tokens or implicitly in hidden or latent representations~\citep{cot, tot, got, coconut, superposition, continuous_cot}. A complementary direction is explored by Recursive Reasoning Models (RRMs), which use repeated computation to refine a persistent latent state rather than to append new elements to an output or reasoning sequence~\citep{looped_transformer, hrm, trm}. This approach is appealing because it decouples reasoning depth from both parameter scale and output length: a compact model can perform many steps of internal computation by repeatedly applying shared transition functions.

Recent recursive reasoning models such as HRM~\citep{hrm} and TRM~\citep{trm} provide early evidence for the potential of this approach in structured reasoning. Rather than producing a solution in a single feedforward pass, they perform extended computation through iterative latent-state refinement, deep supervision across refinement steps, and reasoning-oriented recurrent designs such as hierarchical latent dynamics. These features make them well suited to problems requiring constraint propagation, state tracking, iterative correction, and multi-step inference. More broadly, they build on a principle also explored in recurrent Transformer architectures such as Universal Transformers~\citep{universal} and Looped Transformers~\citep{looped_transformer}: shared Transformer blocks can be repeatedly applied to increase computational depth without increasing parameter count. Together, these models suggest that reasoning capability can emerge not only from scaling model size or generating longer traces, but also from the organization of computation itself.

\begin{figure}
    \centering
\includegraphics[width=0.92\linewidth]{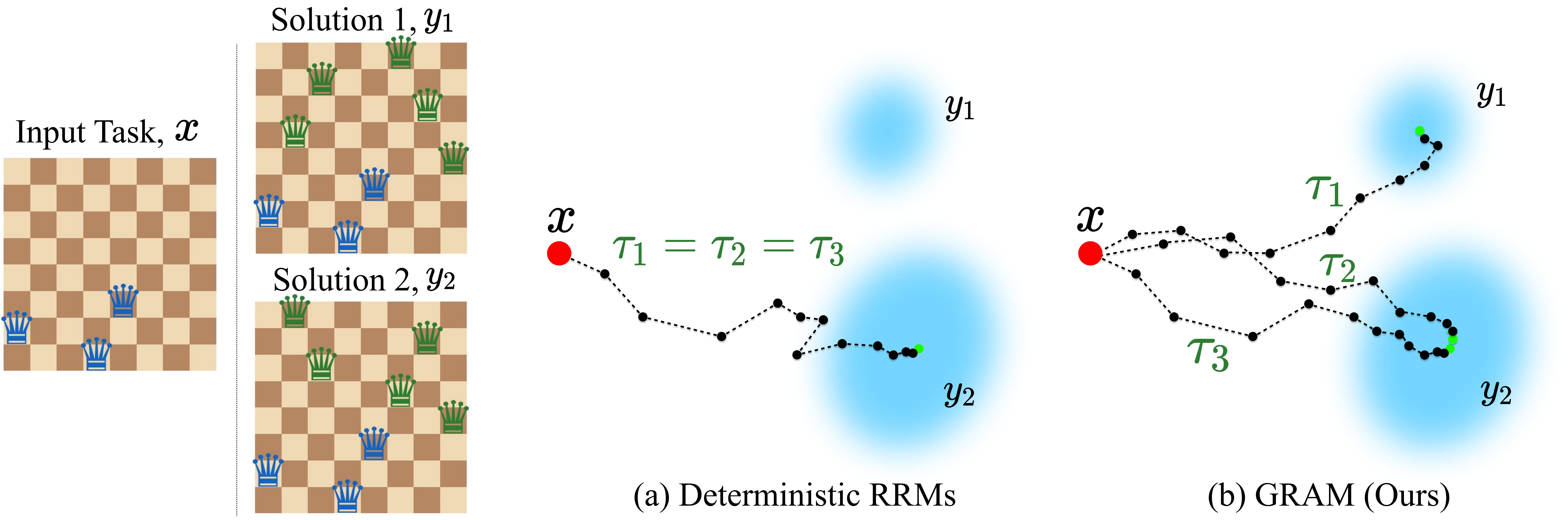}
\caption{\textbf{Comparison of Latent Reasoning Trajectories.} 
Left: N-Queens Example with two valid solutions. Right: Given three independent runs for latent reasoning ($\tau_1, \tau_2, \tau_3$): 
(a) Prior RRMs (e.g. HRM, TRM) are deterministic---all runs collapse to an identical trajectory, converging to a single solution and failing to explore alternatives, while (b) GRAM explores diverse trajectories, producing diverse trajectories that reach multiple valid solutions $y_1$ and $y_2$, while naturally enabling parallel inference-time scaling.}
\label{fig:overview}
\vspace{-2em}
\end{figure}

While recurrent latent-state refinement provides an appealing mechanism for efficiently increasing reasoning depth, depth alone is not sufficient for many reasoning problems. A capable reasoning system should also be able to maintain uncertainty, consider alternative hypotheses, and explore multiple possible solution strategies~\citep{kahneman2011thinking, bengio2017consciousness}. This is especially important in settings where ambiguity or multiple valid solutions are intrinsic, and more generally in problems where a single refinement path may become trapped in a suboptimal reasoning trajectory. In this sense, future RRMs should be not only deep, in the sense of repeated refinement, but also wide, in the sense of maintaining and exploring multiple latent trajectories in parallel. 

Existing RRMs~\citep{looped_transformer,hrm,trm, universal}, however, remain fundamentally deterministic: given the same input and initialization, they follow a single latent trajectory and converge to a single prediction. This deterministic recursion collapses the space of plausible reasoning paths into a single attractor, leaving probabilistic multi-hypothesis latent reasoning largely unexplored within the RRM paradigm. This motivates the central question of our work: \textit{can recursive latent computation support probabilistic, generative, multi-hypothesis reasoning while preserving the efficiency of compact recurrent models?}

In this paper, we propose Generative Recursive reAsoning Models (GRAM), a framework that turns recursive latent reasoning into probabilistic multi-trajectory computation. GRAM treats the reasoning process itself as a stochastic latent trajectory: at each recursion step, the model samples a transition conditioned on the input and the current reasoning state, rather than deterministically updating to a single next state. Repeating this process defines a distribution over possible reasoning trajectories, allowing the model to maintain multiple hypotheses, explore alternative solution strategies, and scale inference not only by increasing recursive depth but also by sampling trajectories in parallel. From a probabilistic perspective, GRAM is a latent-variable generative model. It models $p_\theta(y \mid x)$ by marginalizing over latent reasoning trajectories, while the same recursive process can also define an unconditional generative model $p_\theta(x)$ when the input is fixed or absent. 
% In this way, GRAM recasts recursive reasoning from deterministic iterative refinement into probabilistic generative reasoning over latent trajectories.

We evaluate GRAM on controlled reasoning and generation tasks that serve as probes of the architectural properties targeted by our formulation, namely recursive refinement, stochastic exploration, multi-solution coverage, and inference-time scaling. Sudoku-Extreme~\citep{hrm} and ARC-AGI~\citep{arc, arc2} test structured reasoning under hard constraints and abstract transformations; N-Queens and Graph Coloring evaluate multi-solution reasoning; and Sudoku and binarized MNIST~\citep{mnist} generation from empty inputs investigates the potential of the same recursive process as an unconditional generator. To isolate what stochastic recursion adds, we compare GRAM with the most relevant deterministic recurrent and recursive latent reasoning baselines, including Looped Transformers, HRM, and TRM, rather than frontier-scale general-purpose LLMs whose training data, inference budgets, and external scaffolding are not directly comparable.

Our main contribution is to establish probabilistic multi-trajectory recursion as a design principle for future recurrent and recursive reasoning architectures. Concretely, we make three contributions. First, we formulate recursive reasoning as a latent-variable generative process, where solutions are obtained by marginalizing over stochastic reasoning trajectories. Second, we introduce width-based inference-time scaling, enabling inference to scale not only with recursive depth but also with the number of sampled latent trajectories. Third, we provide empirical evidence that this formulation yields the intended architectural advantages over deterministic recurrent and recursive baselines, improving structured reasoning, multi-solution constraint satisfaction, and unconditional generation.

\section{Generative Recursive Reasoning Models}

In this section, we introduce {Generative Recursive reAsoning Models (GRAM)}, an instantiation of probabilistic recursive reasoning. 
We describe the architecture in \cref{sec:architecture} and the training procedure in \cref{sec:training}, with an architecture schematic shown in \cref{fig:architecture}; pseudocode for both training and inference is given in Appendix~\ref{appx:algorithms}.

\subsection{Architecture}
\label{sec:architecture}
\begin{wrapfigure}{r}{0.5\linewidth}
\vspace{-1em}
\centering
\includegraphics[width=\linewidth]{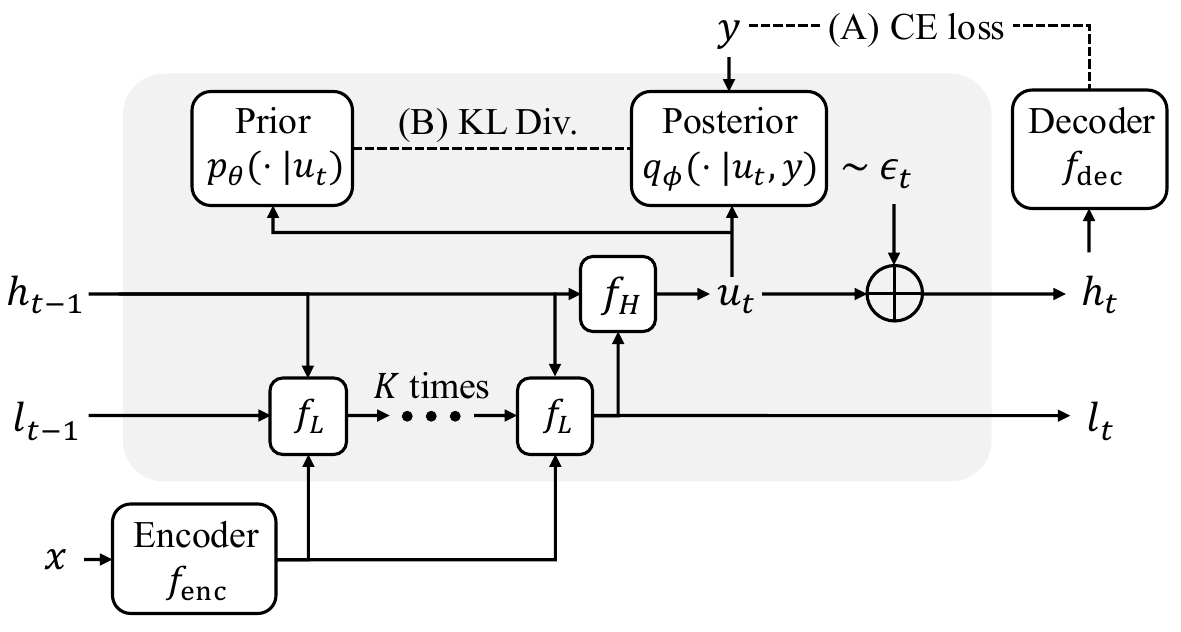}
\caption{\textbf{GRAM Architecture.}
A single stochastic latent transition in the hierarchical instantiation $z=(h,l)$. After $K$ low-level refinements via $f_{\mathrm{L}}$, the high-level update $f_{\mathrm{H}}$ produces a deterministic proposal $u_t$, to which stochastic guidance $\epsilon_t$ is added: $h_t = u_t + \epsilon_t$.}
\vspace{-0em}
\label{fig:architecture}
\end{wrapfigure}
\textbf{Overview.}
GRAM models the conditional distribution $p_\theta(y \mid x)$ by marginalizing over stochastic latent reasoning trajectories. Given an input $x$, GRAM first computes an embedding
\begin{align}
    e_x = f_{\mathrm{enc}}(x;\theta),
\end{align}
which is reused throughout the entire recursive computation. Starting from a fixed initial latent state $z_0$, the model evolves the latent state through learned stochastic transitions. The recursive computation is organized into two nested levels: inner and outer loops.

At the inner level, a \emph{latent transition} samples a new latent state conditioned on the previous latent state and the input embedding,
\begin{align}
    z_{t} \sim p_{\theta}(z_t \mid z_{t-1}, e_x), \qquad t = 1, \dots, T. \label{eq:gram_transition}
\end{align}
At the end of the $T$ transitions, the decoder produces a prediction,
$\hat{y} = \argmax f_{\mathrm{dec}}(z_T;\theta)$.
We refer to the sequence of $T$ transitions from the initial state $z_0$ to the final state $z_T$ as a \emph{supervision step}. A supervision step is the unit at which the decoder is invoked, and the training objective is applied, with gradients computed as described in \cref{sec:training}.

At the outer level, $N_{\mathrm{sup}}$ supervision steps are applied recursively, with the final state of one supervision step serving as the initial state of the next, thereby forming the full recursive computation:
\begin{align}
    z_0^{(1)}
    \;\xrightarrow{\;T\text{ transitions}\;}\;
    z_T^{(1)} = z_0^{(2)}
    \;\xrightarrow{\;T\text{ transitions}\;}\;
    \cdots
    \;\xrightarrow{\;T\text{ transitions}\;}\;
    z_T^{(N_{\mathrm{sup}})},
    \label{eq:supervision_chain}
\end{align}
where $z_t^{(n)}$ denotes the latent state at the $t$-th transition of the $n$-th supervision step, $z_0^{(1)}$ is the fixed initial state, and the terminal state of one supervision step serves as the initial state of the next ($\smash{z_0^{(n+1)} := z_T^{(n)}}$). This abstract formulation can be instantiated with various recurrent Transformer backbones, including flat designs such as Universal Transformers and Looped Transformers~\citep{universal, looped_transformer}, as well as hierarchical designs such as HRM and TRM~\citep{hrm, trm}.

\textbf{Stochastic Latent Transitions.}
Unlike prior recursive reasoning models (RRMs) that update the latent state deterministically and follow a single fixed trajectory~\citep{hrm, trm}, GRAM defines $p_\theta(z_t \mid z_{t-1}, e_x)$ as a stochastic transition, so that repeated computation induces a distribution over latent reasoning trajectories. Concretely, GRAM realizes this transition as a learned stochastic residual perturbation around a deterministic update. At each transition, the model first computes a deterministic update $u_t$ from $z_{t-1}$ and $e_x$, then samples a conditional perturbation from a state-dependent Gaussian, and adds it to $u_t$:
\begin{align}
    \epsilon_t &\sim p_\theta(\epsilon_t \mid u_t) := \mathcal{N}\!\left(\mu_\theta(u_t),\, \sigma_\theta^2(u_t) I\right), \label{eq:gaussian_guidance} \\
    z_t &= u_t + \epsilon_t.
\end{align}
We refer to $\epsilon_t$ as the \textbf{learnable stochastic guidance}. The mean $\mu_\theta(u_t)$ encodes a state-dependent direction in which the trajectory is steered, while the variance $\sigma_\theta^2(u_t)$ controls the amount of exploration. This design allows GRAM to capture uncertainty, prevent convergence to local minima, and support robust exploration of the solution space without discarding the deterministic refinement of $u_t$.

\textbf{Hierarchical Instantiation.} We instantiate the latent state with two interacting components, $z = (h, l)$. The high-level component $h$ is updated once per latent transition and carries abstract reasoning state, while the low-level component $l$ is updated $K$ times within a single transition and carries fine-grained intermediate computation. This decomposition separates the two roles across time scales, with $h$ accumulating slowly across transitions and $l$ refined rapidly within each one.

With this hierarchical multi-scale structure, a single transition $z_{t-1} \to z_t$ is computed as follows. The low-level component is first refined for $K$ updates, with the high-level component held fixed:
\begin{align}
    l_{t,k} = f_{\mathrm{L}}(h_{t-1},\, l_{t,k-1},\, e_x;\, \theta),
    \qquad k = 1, \dots, K,
    \label{eq:low_level}
\end{align}
where $l_{t,0} := l_{t-1}$ and we write $l_t := l_{t,K}$ for the refined low-level component. The high-level component is then updated as a stochastic transition conditioned on the refined $l_t$,
\begin{align}
    u_t       &= f_{\mathrm{H}}(h_{t-1},\, l_t;\, \theta),                                                          \label{eq:deterministic_update} \\
    \epsilon_t &\sim p_\theta(\epsilon_t \mid u_t) := \mathcal{N}\!\big(\mu_\theta(u_t),\, \sigma^2_\theta(u_t)\,I\big), \label{eq:gaussian_prior} \\
    h_t       &= u_t + \epsilon_t,                                                                                  \label{eq:high_level}
\end{align}
and we set $z_t = (h_t, l_t)$. Note that stochasticity is introduced only at the high level: the low-level refinement is fully deterministic, while the stochastic guidance signal $\epsilon_t$ acts on the slower, more abstract component of the latent state, where it can steer the overall reasoning trajectory across transitions. Under this instantiation, the decoder reads only the high-level component, i.e., $f_{\mathrm{dec}}(z_T) = f_{\mathrm{dec}}(h_T)$. Architectural details are in Appendix~\ref{appx:arch_detail}, and Appendix~\ref{app:noise-placement} ablates the level at which stochasticity is injected.

\textbf{Modeling Unconditional Distribution.} While the description so far focuses on the conditional setting $p_\theta(y \mid x)$, the same recursive process can also be defined as an unconditional generative model $p_\theta(x)$ when the input is replaced with an empty conditioning embedding. We use this formulation for unconditional generation tasks in \cref{sec:generation}.

\subsection{Training}
\label{sec:training}

GRAM is trained to model the conditional distribution $p_\theta(y \mid x)$, where each training example consists of an input $x$ and its corresponding target $y$. As a probabilistic model, GRAM adopts a latent-variable formulation and is optimized by maximizing an evidence lower bound (ELBO) with respect to the generative parameters $\theta$ and variational parameters $\phi$.

\textbf{Latent Variable Modeling.}
We model GRAM as a latent-variable probabilistic model $p_{\theta}$, where the full latent trajectory
$\tau = (z_0 \rightarrow \cdots \rightarrow z_{T_{\mathrm{Total}}})$ consists of a sequence of latent variables, with $T_{\mathrm{Total}} = T \times N_{\mathrm{sup}}$.
The conditional likelihood is defined as
\begin{equation}
    p_{\theta}(y \mid x) = \int p_{\theta}(y \mid \tau, x)\, p_{\theta}(\tau \mid x)\, d\tau\,,
\end{equation}
where $x$ denotes the input problem and $y$ denotes the corresponding ground-truth output.

Direct maximum likelihood estimation of $\log p_{\theta}(y \mid x)$ is intractable due to the marginalization over latent trajectories.
We therefore introduce a variational posterior $q_{\phi}(\tau \mid x,y)$ and optimize the evidence lower bound (ELBO), jointly training $\theta$ and $\phi$ via variational inference:
\begin{align}
\log p_{\theta}(y \mid x)
\ge
\mathbb{E}_{q_{\phi}(\tau \mid x,y)}[\log p_{\theta}(y \mid \tau, x)]
-
\mathrm{KL}\!\left(
q_{\phi}(\tau \mid x,y)
\,\|\, 
p_{\theta}(\tau \mid x)
\right).
\end{align}
During training, latent trajectories are sampled from the variational posterior
$q_{\phi}(\cdot \mid x,y)$, which has access to both the input problem $x$ and the
target output $y$. At inference time, where $y$ is unavailable, trajectories are
instead generated from the learned prior $p_{\theta}(\cdot \mid x)$.

Both the prior and the posterior are modeled as conditional Markov processes over latent states:
\begin{equation}
    p_{\theta}(\tau \mid x)
    =
    p(z_0)\prod_{t=1}^{T_{\mathrm{Total}}} p_{\theta}(z_t \mid z_{t-1}, x)\,,
    \qquad
    q_{\phi}(\tau \mid x,y)
    =
    p(z_0)\prod_{t=1}^{T_{\mathrm{Total}}} q_{\phi}(z_t \mid z_{t-1}, x,y)\,.
\end{equation}

Here, $z_0$ is a fixed initial state shared by the prior and posterior. Both transitions are implemented by adding reparameterized Gaussian noise $\epsilon_t$ after a deterministic update $u_t$. The prior and the posterior differ only in the noise head applied on top of $u_t$, and the two heads have identical architecture but separate weights. The prior head parameterizes $p_\theta(\epsilon_t \mid u_t)$ as in \cref{eq:gaussian_prior}; the posterior head parameterizes $q_\phi(\epsilon_t \mid u_t, y) = \mathcal{N}\big(\mu_\phi(u_t, y),\, \sigma^2_\phi(u_t, y) I\big)$. Implementation details of both heads are given in Appendix~\ref{appx:arch_detail}.

Since the two processes share the same Markov structure and all stochasticity is
introduced through $\epsilon_{1:T_{\mathrm{Total}}}$, their trajectory distributions
can be equivalently represented in noise space. Moreover, since GRAM decodes the
output only from the terminal latent state, the likelihood term satisfies
$p_{\theta}(y \mid \tau,x)=p_{\theta}(y \mid z_{T_{\mathrm{Total}}},x)$.
Therefore, the full trajectory-level ELBO can be written as
\begin{align}
\mathcal{L}_{\mathrm{ELBO}}
=
\mathbb{E}_{q_{\phi}}
\big[
    \log p_{\theta}(y \mid z_{T_{\mathrm{Total}}}, x)
\big]
-
\sum_{t=1}^{T_{\mathrm{Total}}}
\mathbb{E}_{q_{\phi}(\epsilon_{<t} \mid x,y)}
\Big[
\mathrm{KL}\!\left(
q_{\phi}(\epsilon_t \mid u_t, y)
\,\|\, 
p_{\theta}(\epsilon_t \mid u_t)
\right)
\Big].
\label{eq:traj_kl_eps}
\end{align}
Here, $u_t = f_{\mathrm{H}}(h_{t-1}, l_t)$ denotes the deterministic high-level update before noise injection, as defined in \cref{eq:deterministic_update}. Since $u_t$ depends on $h_{t-1}$, which is determined by the previously sampled noise variables $\epsilon_{<t}:=(\epsilon_1, \dots, \epsilon_{t-1})$, the expectation averages over these ancestral samples.

\textbf{Practical Implementation.}
In practice, following previous recursive reasoning models~\citep{hrm,trm},
we train GRAM with deep supervision over $N_{\mathrm{sup}}$ consecutive supervision
steps, each consisting of $T$ recursive latent transitions. This provides dense
learning signals along the full latent trajectory, rather than supervising only
the final state after $T_{\mathrm{Total}}=T\times N_{\mathrm{sup}}$ transitions. The
terminal state of each step is reused as the initial state of the next step.

Following standard practice for recurrent models with long computation chains, we apply truncated gradient propagation~\citep{williams1990efficient,unbiasing_bptt}, as used in recent recursive reasoning models~\citep{hrm,trm,scaling_recurrent_depth}. In our implementation, gradients are propagated only through the final transition of each supervision step, $\smash{z_{T-1}^{(n)}\to z_T^{(n)}}$. This gives the following surrogate objective for each supervision step:
\begin{align}
{\mathcal{L}}_{\mathrm{GRAM}}^{(n)}(x,y ; \theta,\phi) =
    \mathbb{E}_{q_{\phi}}
    \big[
        \log p_{\theta}(y \mid z_T^{(n)}, x)
    \big] -
    \mathrm{KL}\big(
        q_{\phi}(\epsilon_T^{(n)} \mid u_T^{(n)}, y)
        \,\|\,
        p_{\theta}(\epsilon_T^{(n)} \mid u_T^{(n)})
    \big)\,,
\end{align}
where $z_T^{(n)}$ is the terminal state of the current supervision step $n$, and gradients are stopped through preceding states. Thus, $\mathcal{L}_{\mathrm{GRAM}}$ should be viewed as a truncated surrogate objective, a biased but memory-efficient approximation to the exact ELBO. Further analysis of this approximation and the training hyperparameters are provided in Appendix~\ref{appx:approx_analysis} and~\ref{appx:train_detail}, respectively. We additionally train an auxiliary halting head alongside the main objective, described in Appendix~\ref{appx:act}.

\subsection{Inference-Time Scaling} \label{sec:inference}
 
GRAM supports two complementary axes of inference-time scaling: \emph{depth}, by varying the number of recursive transitions, and \emph{width}, by sampling multiple latent reasoning trajectories in parallel. For depth, we follow prior recursive reasoning models~\citep{hrm, trm}, where increasing the number of transitions at inference continues to improve performance, as shown in \cref{fig:scaling}. For width, which the stochastic transitions make available, we draw $\smash{\{\tau^{(i)}\}_{i=1}^{N} \sim p_\theta(\tau \mid x)}$ from the learned prior and decode each terminal state into a candidate output $\smash{\hat{y}^{(i)} = f_{\mathrm{dec}}(z_T^{(i)})}$, exploring multiple stochastic reasoning paths simultaneously rather than extending a single trajectory.

To select among candidates, we use majority voting~\citep{selfconsistency}, which returns the most frequent prediction and adds no parameters or training, so the reported gains are attributable to parallel sampling itself. This improves robustness and solution quality without increasing the sequential recursion length, and the complete procedure is summarized in \cref{alg:inference}. Other selectors are possible, such as the Latent Process Reward Model (LPRM) described in Appendix~\ref{appx:lprm}.

\section{Related Work}

\textbf{Latent Reasoning.}
Latent reasoning aims to reduce the inefficiency and verbosity of explicit Chain-of-Thought (CoT) by shifting part or all of the reasoning process into latent or continuous representations~\citep{cot, tot, got, coconut, superposition, continuous_cot}. By avoiding token-by-token generation of intermediate steps, such representations can make reasoning traces more compact and reduce generation overhead. Existing approaches instantiate this idea through hidden states, latent or soft tokens, continuous thoughts, internal reasoning traces, and recursive state updates for scaling test-time computation~\citep{coconut, looped_transformer, mixture_of_inputs, softthinking, softtokens, codi, hybrid_latent_reasoning, scaling_recurrent_depth, mixture_of_recursion, cotformer, relaxed_recursive}. However, many remain organized around autoregressive sequence generation, where additional computation is tied to generating more tokens, latent positions, or sequential reasoning states.

\textbf{Recursive Architectures.}
Recursive architectures perform iterative state updates and have evolved from RNNs to weight-sharing Transformers with adaptive computation~\citep{looped_transformer, universal, rnn, lstm, gru, albert, depth_adaptive, adaptive, cotformer}. Recent recursive reasoning models show that increasing inference-time depth can outperform larger static models~\citep{hrm, trm, scaling_recurrent_depth, mixture_of_recursion}. GRAM builds on this line but formulates recurrence as a probabilistic process: instead of following a single deterministic refinement path, it maintains stochastic latent trajectories, enabling multi-path exploration and generative sampling.

\textbf{Probabilistic Latent State-Space Models.}
Probabilistic recurrent models use stochastic latent transitions to capture uncertainty and multimodal dynamics, often trained with variational inference~\citep{vrnn, srnn, deep_kalman_filter, rssm, dreamerv2, dreamerv3}. They have been widely used in sequential generative modeling, video prediction, and model-based reinforcement learning. GRAM shares this latent state-space view but reinterprets stochastic dynamics as computation rather than temporal observation modeling: latent transitions define possible reasoning trajectories, supporting multi-hypothesis exploration and both conditional $p_\theta(y \mid x)$ and unconditional $p_\theta(x)$ generation.

\section{Experiments}
\label{sec:experiments}

We evaluate GRAM against deterministic recursive baselines on tasks designed to probe what stochastic recursion changes. Stochastic transitions can benefit reasoning in two ways: during training, they expose the model to multiple trajectories for the same input, acting as a form of data augmentation; at inference, they allow the model to explore multiple trajectories and select among them. We therefore report both single-sample ($N{=}1$) and multi-sample ($N{=}20$) results. The $N{=}1$ setting isolates the training-time benefit by comparing all models on equal terms, while $N{=}20$ additionally captures the benefit of inference-time search, which deterministic models cannot exploit.

% We first evaluate structured reasoning on Sudoku-Extreme and ARC-AGI, where each input has a single correct answer (Section~\ref{sec:puzzle}). We then assess multi-solution behavior on N-Queens and Graph Coloring, where an input can have multiple valid solutions (Section~\ref{sec:csp}). Next, we examine unconditional generation of Sudoku boards and binarized MNIST, where no input is given (Section~\ref{sec:generation}). Finally, we ablate the components of GRAM's architecture and stochastic transition (Section~\ref{sec:ablation}).

\begin{figure*}
    \centering
    \includegraphics[width=\linewidth]{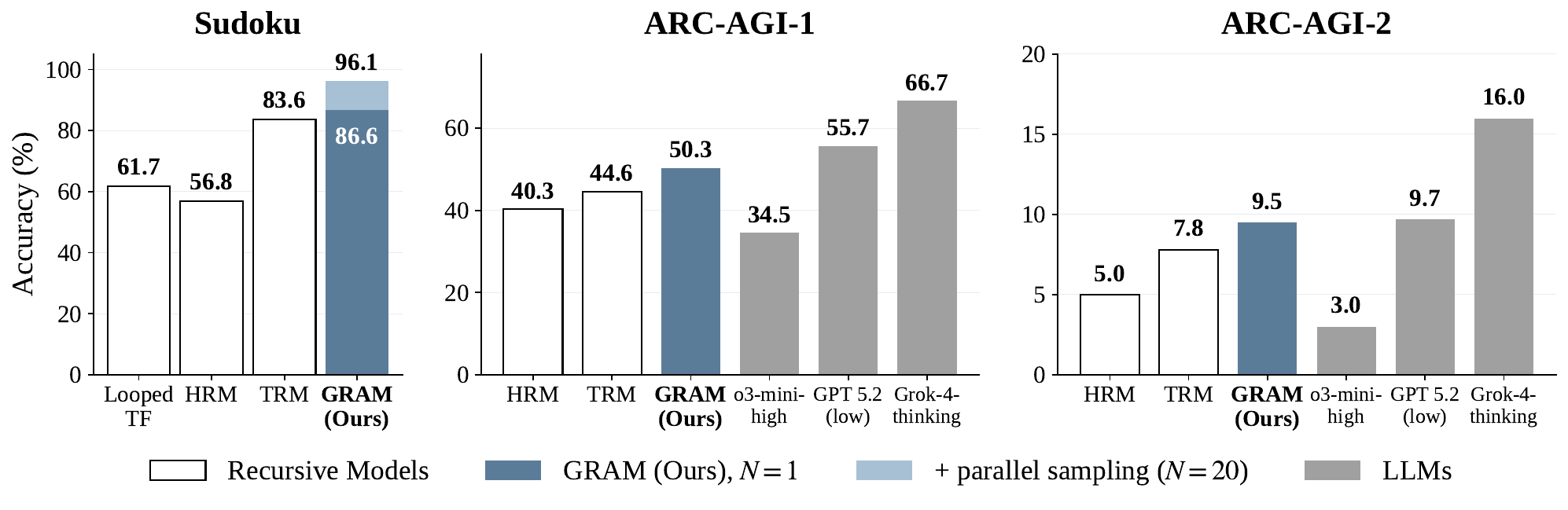}
    % \vspace{-0.5em}
    \caption{\textbf{Performance on puzzle benchmarks.} GRAM outperforms all deterministic recursive baselines on both Sudoku-Extreme and ARC-AGI even without parallel sampling. On Sudoku-Extreme, the GRAM bar is decomposed into its single-sample accuracy ($N{=}1$) and the further gain from parallel sampling ($N{=}20$). The ARC-AGI bars are reported without parallel sampling, since the standard pass@2 protocol (Appendix~\ref{appx:arc_protocol}) already aggregates predictions over 1k augmented inputs. Without test-time augmentation, parallel sampling still improves GRAM on ARC-AGI (Appendix~\ref{app:scaling}). Large reasoning model scores are external reference points for benchmark difficulty, and full results are in \cref{tab:puzzle}.}
    \vspace{-1em}
    \label{fig:puzzle_task}
\end{figure*}

\subsection{Challenging Puzzle Tasks}
\label{sec:puzzle}

\begin{figure}[t]
    \centering
    \includegraphics[width=\linewidth]{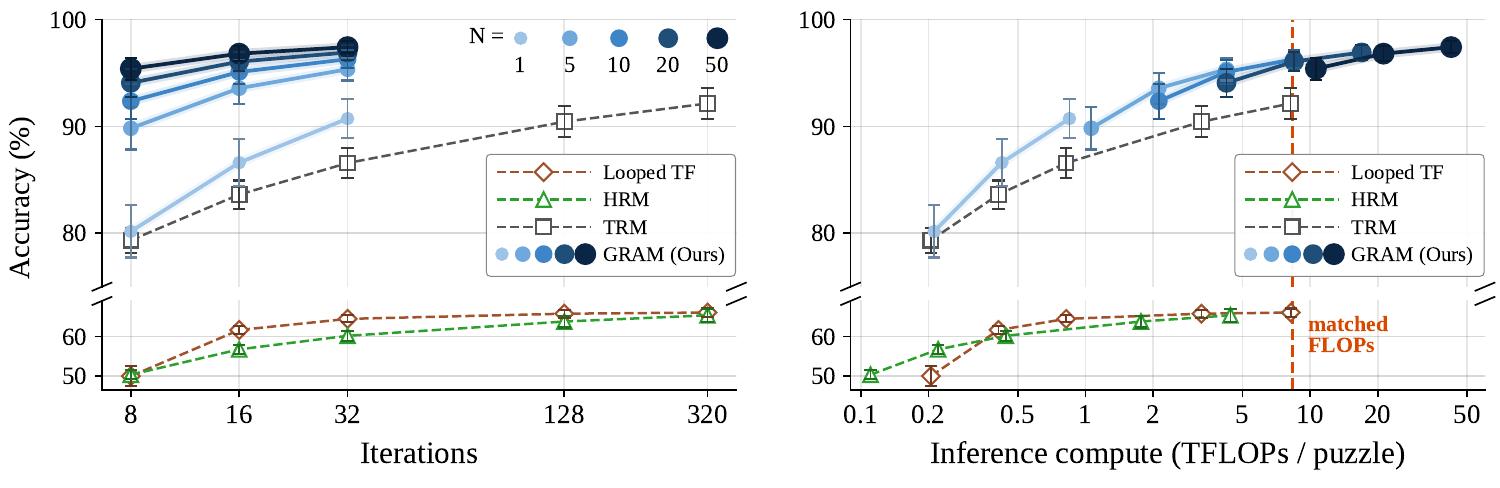}
\caption{%
    \textbf{Inference-time scaling on Sudoku-Extreme.}
    \textbf{(Left) Recursion depth and parallel sampling.} While all models benefit from longer recursion (x-axis), GRAM additionally scales with parallel sampling ($N$ = number of samples). Each iteration corresponds to a supervision step, which amounts to $K\times$ more flat iterations for Looped TF.
    \textbf{(Right) Accuracy versus inference compute.} In FLOPs analysis, GRAM is more accurate than all baselines at every budget. The dashed line marks equal compute for TRM at 320 iters and GRAM at 16 iters with $N{=}20$.
    Curves show the mean over 3 seeds with one standard deviation.
}
    \label{fig:scaling}
    \vspace{-1em}
\end{figure}
\textbf{Setup.} We evaluate on Sudoku-Extreme~\citep{hrm}, which contains 9$\times$9 puzzles with minimal clues requiring extensive constraint propagation, and ARC-AGI Challenge~\citep{arc,arc2}, which tests abstract visual reasoning through few-shot pattern recognition. For Sudoku-Extreme, we report GRAM with a single sample and with $N{=}20$ parallel samples aggregated by majority voting. For ARC-AGI, we report results without parallel sampling, since the standard pass@2 protocol (Appendix~\ref{appx:arc_protocol}) already aggregates predictions over 1,000 augmented inputs. We compare against direct prediction (Transformer) and recursive baselines (Looped TF, HRM, TRM). Looped TF is evaluated on Sudoku-Extreme only, as training it on ARC-AGI is prohibitively expensive. On Sudoku-Extreme, the published results of TRM~\citep{trm} (87.4\%) and HRM~\citep{hrm} (55.0\%) are single runs, which neither capture variance across seeds nor are directly comparable to our 3-seed means. We therefore reproduce all recursive baselines, including those in the scaling analysis, with their official code and configurations and report the mean over 3 seeds. Large reasoning model results~\citep{arcprize2026arcagi_benchmarking} are included as external reference points for benchmark difficulty, not as controlled baselines.

\textbf{Probabilistic Recursion Improves Structured Reasoning.} With a single sample, GRAM outperforms all recursive baselines on both benchmarks (Figure~\ref{fig:puzzle_task}), reaching 86.6\% against TRM's 83.6\% on Sudoku-Extreme, 50.3\% against 44.6\% on ARC-AGI-1, and 9.5\% against 7.8\% on ARC-AGI-2. Parallel sampling lifts GRAM further to 96.1\% on Sudoku-Extreme. We attribute this single-sample lead to training, where stochastic transitions expose GRAM to diverse intermediate states rather than a single fixed path. Full results, including large reasoning models, are given in Table~\ref{tab:puzzle} (Appendix~\ref{app:puzzle_analysis}).

\textbf{Parallel Sampling Provides a New Inference-Time Scaling Axis.} Parallel sampling adds a scaling axis that deterministic models lack. As Figure~\ref{fig:scaling}~(left) shows, more samples improve GRAM on Sudoku-Extreme at every iteration count, and at 16 iterations accuracy rises from 86.6\% at $N{=}1$ to 93.6\% at $N{=}5$ and 96.1\% at $N{=}20$. Drawing 20 samples multiplies the compute accordingly, yet GRAM uses this compute more effectively than deeper recursion, reaching 96.1\% with 8.48 TFLOPs per puzzle, whereas TRM given nearly the same compute through 320 iterations reaches 92.1\% with 8.19 TFLOPs (Figure~\ref{fig:scaling}, right). Moreover, because the samples run in parallel, this compute takes far less time, 184\,ms per puzzle against 2,755\,ms for TRM, whose recursion steps must run sequentially. Appendix~\ref{appx:flops} reports the compute and latency measurements in detail.

\subsection{Multi-solution Puzzle Tasks}
\label{sec:csp}

\textbf{Setup.} To evaluate whether GRAM can solve problems whose inputs have multiple valid solutions, we test on N-Queens ($8 \times 8$, $10 \times 10$) and Graph Coloring (8-vertex, 10-vertex). We compare against direct prediction (Transformer), recursive models (Looped TF, HRM, TRM), and generative models (MDLM~\citep{mdlm} and an autoregressive Transformer, denoted AR).
We report accuracy, whether the output satisfies all constraints, for N-Queens, and conflict edges, the number of constraint violations, for Graph Coloring. Both are reported in Table~\ref{tab:csp} from a single output of every model, and also with 20 samples for AR, MDLM, and GRAM. Coverage, the fraction of valid solutions recovered, is measured with 20 samples. Appendix~\ref{sec:multi_solution_details} describes how both datasets are constructed.

\textbf{Deterministic Recursion Fails on Multi-Solution Tasks.} \Cref{tab:csp} shows that deterministic recursive models fall behind in both accuracy and coverage on every multi-solution task. On $8\times8$ N-Queens, they reach at most 78.7\% accuracy against 99.5\% for GRAM and recover at most 26.7\% of valid solutions, whereas sampling models recover 84.8\% or more. The gap widens as inputs admit more solutions: from $8\times8$ to $10\times10$, where the average number of solutions per input rises from 1.65 to 2.93, the accuracy gap between TRM and GRAM grows from 32.7 to 72.2 points and deterministic coverage drops to at most 6.2\%. Graph Coloring shows the same trend from 8 to 10 vertices, and grouping N-Queens inputs by solution count confirms it at the instance level: deterministic models lose accuracy as solutions increase, while GRAM stays flat even with a single sample (Appendix~\ref{appx:coverage}). Inputs with more valid solutions carry more conflicting targets, which a deterministic model must collapse into a single output, whereas GRAM absorbs this ambiguity in its latent trajectory.

\textbf{Recursive Refinement Yields Sharper Constraint Satisfaction.} Generative models (AR, MDLM) already achieve high coverage, but GRAM satisfies constraints more reliably in most settings while attaining the highest coverage in all four. It reaches 99.5\% accuracy on $8\times8$ N-Queens and leaves under a third of AR's conflict edges on Graph Coloring. Even when all generative models draw 20 samples, GRAM still exceeds them on most tasks (\cref{tab:csp}). Together, these results suggest that recursive refinement enables stricter constraint satisfaction than generative sampling alone.

\begin{table*}[t]
\caption{\textbf{Evaluation on N-Queens and Graph Coloring benchmarks.} Rec. and Gen. indicate whether the model uses recursive computation and generative sampling, respectively. Values are mean $\pm$ standard deviation over runs, with accuracy and coverage in percent. $N$ is the number of samples for accuracy and conflicts.}
\label{tab:csp}
\centering
\Large
\providecommand{\cspstd}[1]{{\scriptsize$\pm#1$}}
\adjustbox{max width=\textwidth}{
\setlength{\heavyrulewidth}{1.5pt}
\begin{tabular}{lccccccccccccc}
\toprule
& & & & & \multicolumn{4}{c}{\textbf{N-Queens}} & \multicolumn{4}{c}{\textbf{Graph Coloring}} \\
\cmidrule(lr){6-9} \cmidrule(lr){10-13}
& & & & & \multicolumn{2}{c}{$8 \times 8$} & \multicolumn{2}{c}{$10 \times 10$} & \multicolumn{2}{c}{8-vertex} & \multicolumn{2}{c}{10-vertex} \\
\cmidrule(lr){6-7} \cmidrule(lr){8-9} \cmidrule(lr){10-11} \cmidrule(lr){12-13}
Method & Rec. & Gen. & \# Params & $N$ & Acc. & Cov@20 & Acc. & Cov@20 & Conflict$\downarrow$ & Cov@20 & Conflict$\downarrow$ & Cov@20 \\
\midrule
Direct Pred (8 layers) & \ding{55} & \ding{55} & 27M & -- & 40.4\cspstd{1.1} & 13.7\cspstd{1.1} & 13.6\cspstd{0.5} & 1.6\cspstd{0.2} & 179.3\cspstd{4.0} & 19.9\cspstd{0.2} & 198.7\cspstd{5.0} & 6.7\cspstd{0.1} \\
Direct Pred (32 layers) & \ding{55} & \ding{55} & 100M & -- & 40.2\cspstd{1.3} & 13.6\cspstd{1.1} & 13.1\cspstd{0.4} & 1.6\cspstd{0.2} & 174.0\cspstd{18.0} & 19.1\cspstd{1.7} & 227.7\cspstd{34.5} & 6.5\cspstd{1.9} \\
\midrule
Looped TF & \checkmark & \ding{55} & 7M & -- & 68.1\cspstd{7.4} & 23.6\cspstd{1.9} & 50.0\cspstd{7.6} & 6.2\cspstd{3.2} & 136.0\cspstd{16.1} & 20.5\cspstd{1.5} & 157.3\cspstd{9.0} & 7.2\cspstd{0.7} \\
HRM & \checkmark & \ding{55} & 27M & -- & 78.7\cspstd{2.9} & 26.7\cspstd{1.3} & 37.8\cspstd{0.7} & 4.7\cspstd{0.1} & 109.7\cspstd{1.5} & 21.8\cspstd{0.3} & 164.3\cspstd{21.6} & 8.9\cspstd{1.7} \\
TRM & \checkmark & \ding{55} & 7M & -- & 66.8\cspstd{5.6} & 22.6\cspstd{0.8} & 17.5\cspstd{11.2} & 2.0\cspstd{1.3} & 109.3\cspstd{3.1} & 22.3\cspstd{0.6} & 170.7\cspstd{17.9} & 6.8\cspstd{0.3} \\
\midrule
AR & \ding{55} & \checkmark & 10.6M & 1 & 96.3\cspstd{1.0} & -- & \textbf{90.0}\cspstd{2.2} & -- & 37.3\cspstd{6.1} & -- & 101.7\cspstd{4.0} & -- \\
& & & & 20 & 98.3\cspstd{0.7} & 84.8\cspstd{0.8} & 95.1\cspstd{1.5} & 53.2\cspstd{0.8} & 19.0\cspstd{11.3} & 83.0\cspstd{0.7} & 61.3\cspstd{8.3} & 40.0\cspstd{0.3} \\
\addlinespace[2pt]
MDLM & \ding{55} & \checkmark & 12.6M & 1 & 96.1\cspstd{1.5} & -- & 74.3\cspstd{6.6} & -- & 16.3\cspstd{2.5} & -- & 57.3\cspstd{4.7} & -- \\
& & & & 20 & \textbf{100.0}\cspstd{0.0} & 87.2\cspstd{0.6} & 98.6\cspstd{0.8} & 47.4\cspstd{2.2} & \textbf{2.7}\cspstd{0.6} & 84.5\cspstd{4.0} & 12.0\cspstd{7.0} & 48.2\cspstd{1.4} \\
\midrule
\rowcolor{blue!8}
\textbf{GRAM (Ours)} & \checkmark & \checkmark & 10M & 1 & \textbf{99.5}\cspstd{0.3} & -- & 89.7\cspstd{2.7} & -- & \textbf{8.3}\cspstd{2.5} & -- & \textbf{31.3}\cspstd{15.0} & -- \\
\rowcolor{blue!8}
& & & & 20 & \textbf{100.0}\cspstd{0.0} & \textbf{89.7}\cspstd{1.7} & \textbf{100.0}\cspstd{0.0} & \textbf{57.5}\cspstd{3.4} & \textbf{2.7}\cspstd{2.1} & \textbf{85.8}\cspstd{0.5} & \textbf{3.3}\cspstd{1.5} & \textbf{51.3}\cspstd{2.8} \\
\bottomrule
\end{tabular}
}
\vspace{-1em}
\end{table*}

\subsection{Exploring GRAM as an Unconditional Generator}
\label{sec:generation}

\begin{wrapfigure}{r}{0.45\textwidth}
\vspace{-5mm}
\centering

% ---- Table ----
\captionof{table}{\textbf{Unconditional generation results on binarized MNIST.} We report IS ($\uparrow$) and FID ($\downarrow$). For iterative models, a step corresponds to a supervision step for TRM and GRAM, and a denoising step for D3PM. FID is calculated using real samples with original pixel values (0--255).}
\label{tab:generation}

\scriptsize
\setlength{\heavyrulewidth}{1.5pt}
\begin{tabular}{@{}lcc@{}}
\toprule
Method & IS ($\uparrow$) & FID ($\downarrow$) \\
\midrule
VAE & 1.70 & 86.28 \\
D3PM (1000 steps)  & 1.86 & 74.03 \\
TRM (16 steps)     & 1.00 & 303.29 \\
\midrule
\textbf{GRAM (Ours)} \\
\quad 16 steps  & 1.89 & 77.79 \\
\quad 64 steps  & 1.95 & 75.39 \\
\quad 256 steps & \textbf{2.04} & \textbf{73.34} \\
\bottomrule
\end{tabular}

\vspace{2em}

% ---- Figure ----
\includegraphics[width=0.68\linewidth]{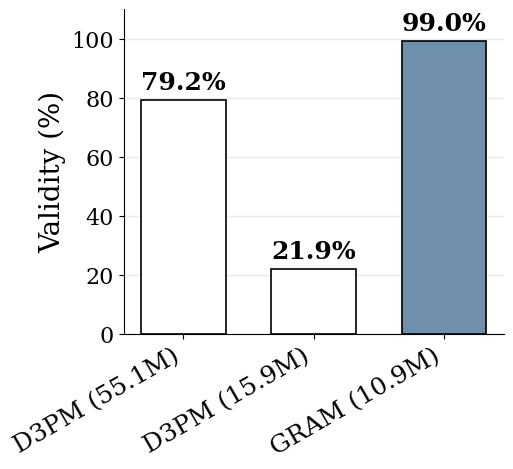}
\captionof{figure}{\textbf{Unconditional Sudoku generation.} Validity (\%) of generated Sudoku puzzles. GRAM achieves higher validity than D3PM with substantially fewer parameters and steps.}
\label{fig:sudoku_generation_plot}

\vspace{-3em}
\end{wrapfigure}

\textbf{Setup.}
To test GRAM's unconditional generative capability, we evaluate it on Sudoku generation from empty boards, measured by the fraction of valid boards, and on binarized MNIST generation, measured by IS~\citep{salimans2016improved} and FID~\citep{fid}. GRAM replaces the input with an empty conditioning signal and samples from its learned prior. We compare with D3PM~\citep{d3pm} on both tasks, and additionally with VAE~\citep{vae} and TRM on MNIST. All models have roughly matched parameter counts; on Sudoku generation task, we also evaluate a D3PM about five times larger to test whether its performance is capacity-limited. Further details are in Appendix~\ref{appx:train_detail}.

\begin{figure}[t]
\centering
\includegraphics[width=\linewidth]{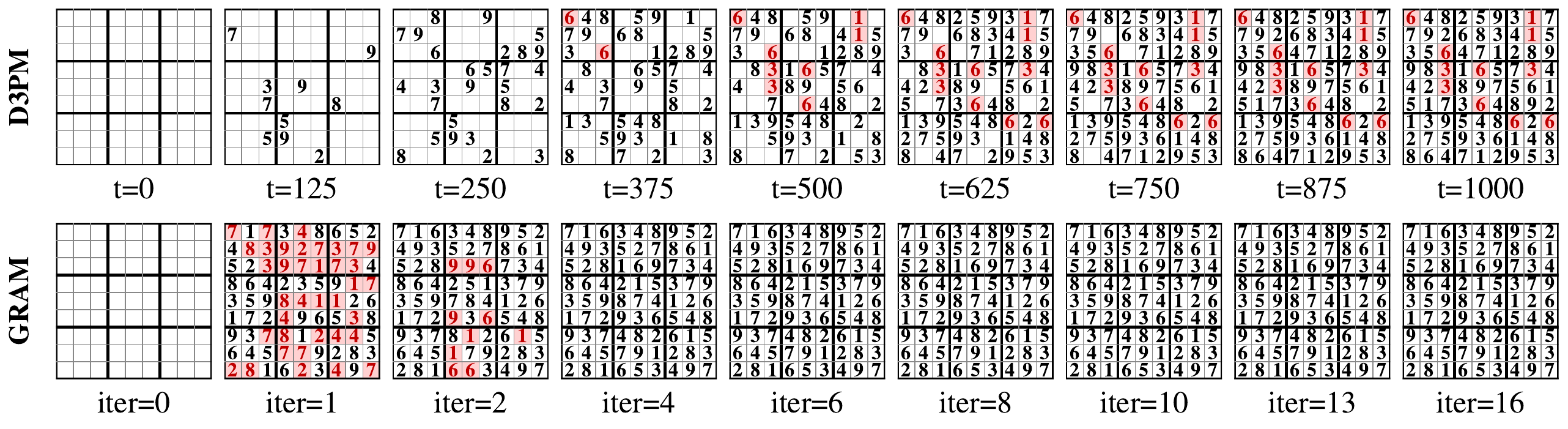}
\caption{\textbf{Generation trajectories of D3PM and GRAM on unconditional Sudoku.} Each row shows successive states of a single generation from an empty grid, D3PM over denoising steps $t$ and GRAM over recursion steps. GRAM proposes a complete board at its first step and corrects it in place, while D3PM fills cells progressively. Incorrect digits are highlighted in red.}
\label{fig:sudoku_gen_vis}
\vspace{-1em}
\end{figure}

\textbf{Probabilistic Recursion Generalizes to Unconditional Generation.} Unconditional generation tests the formulation directly. With no input, the entire output must come from the latent trajectory. Deterministic recursion therefore follows the same trajectory for every sample and collapses to a single output, with TRM producing a near-constant output on binarized MNIST. GRAM instead samples diverse digits from its prior, outperforming VAE with 16 steps and surpassing D3PM on both IS and FID with 256 steps, still far fewer than D3PM's 1,000 steps (\cref{tab:generation}). The gap widens on structured generation, where GRAM produces 99.05\% valid Sudoku boards, all distinct among 100K samples, whereas D3PM reaches at most 79.18\% even at five times GRAM's size (\cref{fig:sudoku_generation_plot}). \Cref{fig:sudoku_gen_vis} contrasts the generation trajectories of GRAM and D3PM, suggesting that refining a complete candidate suits hard-constraint generation better than filling cells progressively.

\begin{figure*}
    \centering
    \includegraphics[width=\linewidth]{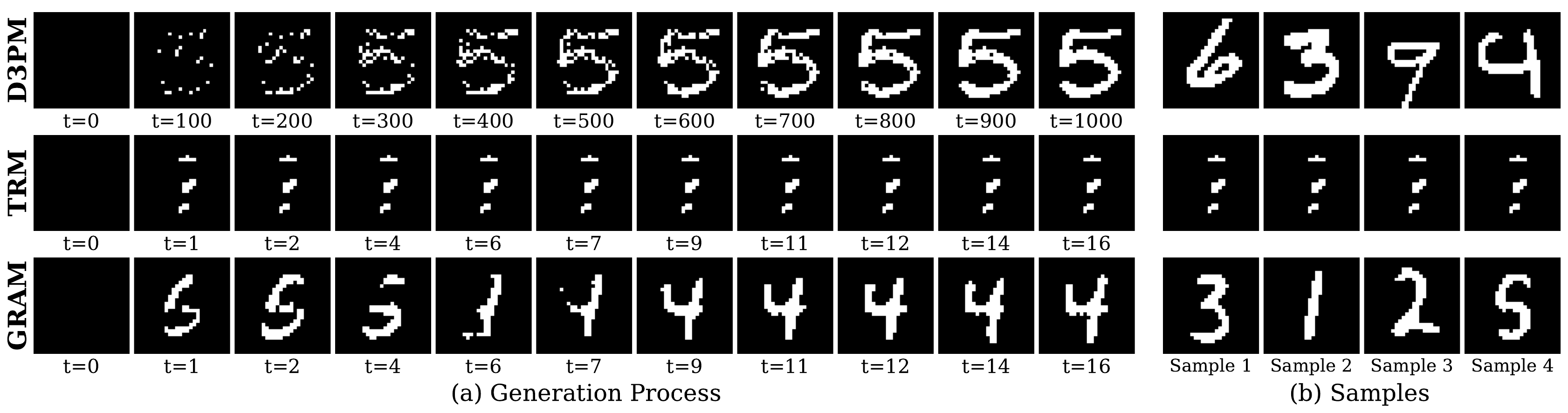}
    \caption{\textbf{Visualization of the generation process and samples.} (a) The generation process over recursion steps. Each row corresponds to a different model. GRAM (bottom) progressively refines the generated image through recursive latent updates, correcting initial errors. (b) Unconditional generated samples from each model.}
    \label{fig:img_generation}
    \vspace{-1em}
\end{figure*}

\textbf{Inference-Time Scaling Transfers to Generation Tasks.}
\Cref{tab:generation} shows that increasing recursive inference steps from 16 to 256 monotonically improves both IS ($1.89 \!\to\! 2.04$) and FID ($77.79 \!\to\! 73.34$), despite training with only 16 supervision steps. This shows that GRAM can use additional recursive computation at inference to improve generation quality. \Cref{fig:img_generation} visualizes the recursive refinement process; additional samples from GRAM and the VAE baseline are provided in Appendix~\ref{app:generation}.

\subsection{Ablation Study}
\label{sec:ablation}

We ablate GRAM's model structure and stochastic mechanism on Sudoku-Extreme and N-Queens ($8\times8$). Table~\ref{tab:ablation} reports the architecture ablation in (a) and the stochasticity ablation in (b).

\begin{table}[t]
\caption{\textbf{Ablation study on Sudoku-Extreme and N-Queens ($8\times8$).} Means and sample standard deviations over three seeds. (a) compares architectural variants built from the Looped TF baseline (DS = deep supervision, HR = hierarchical recursion, SG = stochastic guidance). (b) ablates the stochasticity and guidance mechanisms.}
\label{tab:ablation}
    \centering
    \scriptsize
    \setlength{\tabcolsep}{2pt}
    \renewcommand{\arraystretch}{1.1}
    \begin{subtable}[t]{0.53\linewidth}
        \centering
        \caption{Architecture Ablation.}
        \label{tab:ablation_arch}
        \begin{tabular}{@{}lcccc@{}}
        \toprule
        & \multicolumn{2}{c}{\textbf{Sudoku}} & \multicolumn{2}{c}{\textbf{N-Queens}} \\
        \cmidrule(lr){2-3} \cmidrule(lr){4-5}
        \textbf{Model variant} & $N$=1 & $N$=20 & $N$=1 & $N$=20 \\
        \midrule
        Base (Looped TF) & 61.7$_{\color{gray}\pm0.9}$ & 61.7$_{\color{gray}\pm0.9}$ & 68.1$_{\color{gray}\pm7.4}$ & 68.1$_{\color{gray}\pm7.4}$ \\
        + DS + HR (=TRM) & 83.6$_{\color{gray}\pm1.3}$ & 83.6$_{\color{gray}\pm1.3}$ & 66.8$_{\color{gray}\pm5.6}$ & 66.8$_{\color{gray}\pm5.6}$ \\
        + SG & 65.1$_{\color{gray}\pm0.5}$ & 74.0$_{\color{gray}\pm0.6}$ & 87.0$_{\color{gray}\pm7.5}$ & 87.3$_{\color{gray}\pm7.8}$ \\
        + DS + SG & 76.9$_{\color{gray}\pm2.6}$ & 78.8$_{\color{gray}\pm1.5}$ & 82.6$_{\color{gray}\pm10.7}$ & 83.2$_{\color{gray}\pm10.2}$ \\
        \midrule
        \makecell[l]{\textbf{+ DS + HR + SG}\\\textbf{(=GRAM)}} & 86.6$_{\color{gray}\pm2.2}$ & \textbf{96.1}$_{\color{gray}\pm0.9}$ & 99.5$_{\color{gray}\pm0.3}$ & \textbf{100.0}$_{\color{gray}\pm0.0}$ \\
        \bottomrule
        \end{tabular}
    \end{subtable}
    \hfill
     \begin{subtable}[t]{0.465\linewidth}
        \centering
        \caption{Mechanism Ablation ($N$=20).}
        \label{tab:ablation_mech}
        \begin{tabular}{@{}lcc@{}}
        \toprule
        \textbf{Model variant} & \textbf{Sudoku} & \textbf{N-Queens} \\
        \midrule
        \textbf{GRAM (ours)} & \textbf{96.1}$_{\color{gray}\pm0.9}$ & \textbf{100.0}$_{\color{gray}\pm0.0}$ \\
        \midrule
        w/o stochastic guidance & 83.6$_{\color{gray}\pm1.3}$ & 66.8$_{\color{gray}\pm5.6}$ \\
        stochasticity only & 96.0$_{\color{gray}\pm1.1}$ & 62.1$_{\color{gray}\pm14.9}$ \\
        guide only & 0.0$_{\color{gray}\pm0.0}$ & 0.0$_{\color{gray}\pm0.0}$ \\
        w/ direct prediction & 70.7$_{\color{gray}\pm7.6}$ & 98.8$_{\color{gray}\pm2.2}$ \\
        \midrule
        TRM w/ stochastic decoder & 79.7$_{\color{gray}\pm1.9}$ & 65.9$_{\color{gray}\pm4.7}$ \\
        TRM w/ random init & 91.2$_{\color{gray}\pm0.9}$ & 71.4$_{\color{gray}\pm8.5}$ \\
        \bottomrule
        \end{tabular}
    \end{subtable}
    \vspace{-1em}
\end{table}

\textbf{Stochastic Guidance Provides Consistent Gains Across Architectures.} 
\cref{tab:ablation_arch} shows that stochastic guidance (SG) improves both the flat Looped TF baseline and the hierarchically recursive TRM variant. With SG, Looped TF reaches 74.0\% on Sudoku and 87.3\% on N-Queens at $N=20$. Combined with deep supervision and hierarchical recursion, SG yields the full GRAM model, achieving the best overall performance (96.1\%). These results demonstrate that SG remains effective across different model structures.

\textbf{Stochasticity and Guidance Play Complementary Roles.} \Cref{tab:ablation_mech} ablates each component of the learned distribution $\epsilon_t \sim \mathcal{N}(\mu_\theta, \sigma_\theta^2 I)$ in~\cref{eq:gaussian_guidance}. We report $N{=}20$, where the effect of stochasticity is most visible; single-sample results in Appendix~\ref{app:ablation_n1} lead to the same conclusions. Removing guidance (\textit{stochasticity only}, $\epsilon_t \sim \mathcal{N}(0, \sigma_\theta^2 I)$) retains strong performance on Sudoku (96.0\%) but degrades substantially on N-Queens (62.1\%). Removing stochasticity (\textit{guide only}, $\epsilon_t=\mu_\theta$) fails completely on both tasks (0.0\%). Stochasticity thus provides the exploration that recursion relies on, while guidance steers this exploration toward valid solutions, which matters most when multiple solutions compete as in N-Queens. Beyond these two components, the \textit{w/ direct prediction} variant samples $z_t \sim \mathcal{N}(\mu_\theta(u_t), \sigma_\theta^2(u_t)I)$ directly instead of using the residual update $z_t=u_t+\epsilon_t$, and thus no longer recovers the deterministic update $z_t=u_t$ when $\epsilon_t=0$. Its inconsistent performance across Sudoku (70.7\%) and N-Queens (98.8\%) suggests that separating deterministic recursion from stochastic guidance yields a more robust transition structure.

\textbf{Variational Stochastic Recursion Outperforms Simple Randomization.}
We test two simple approaches to add stochasticity to TRM: (1) \textit{stochastic decoding}, which samples from the output distribution instead of argmax, and (2) \textit{random initialization}, which samples $z_0$ from a Gaussian $\mathcal{N}(0, I)$ at each inference. Stochastic decoding provides no meaningful improvement over deterministic TRM. With majority voting at $N=20$, random initialization improves Sudoku to 91.2\%, but remains below GRAM (96.1\%). On N-Queens, it provides only a modest improvement over TRM (66.8\% to 71.4\%), remaining far below GRAM (100.0\%). These results suggest that variationally modeling stochastic transitions within the recursive process is important for achieving consistent gains.

\section{Conclusions and Limitations}
We introduced GRAM, a generative framework that transforms deterministic recursive architectures into probabilistic generative models capable of modeling both $p(y \mid x)$ and $p(x)$ via recursive amortized variational inference. For reasoning problems, introducing stochasticity into latent transitions enables diverse solution discovery and improved exploration compared to deterministic counterparts. Notably, we demonstrate GRAM can leverage width-based inference-time scaling as a complement to depth, by sampling multiple latent trajectories in parallel and bypassing the latency bottleneck of depth-only scaling. Our ablations further reveal that stochastic guidance is a general-purpose extension that improves recursion whether flat or hierarchical, and that the gains stem specifically from the variational framework --- not from mere randomness, as naive stochastic alternatives applied to existing models fall well short of GRAM. Beyond solution-seeking, GRAM also demonstrates potential as an unconditional generative model through recursion-based generation over inputs, with generation quality improving monotonically with recursive depth even beyond training-time steps.

These results suggest new directions for generative modeling via hierarchical recursion. Despite these strengths, the sequential nature of deep supervision limits training efficiency compared to Transformers, posing a significant barrier to scaling GRAM toward larger foundation models.

\newpage
\section*{Acknowledgment}
This research was supported by the GRDC (Global Research Development Center) Cooperative Hub Program (RS-2024-00436165) through the National Research Foundation of Korea (NRF) funded by the Ministry of Science and ICT (MSIT) and by the Institute of Information \& Communications Technology Planning \& Evaluation (IITP) grants funded by the Korea government (MSIT) (No. RS-2024-00509279, Global AI Frontier Lab, NYU-KAIST Global Innovation and Research Institute; No. RS-2026-25526850, HERMES-Net; and No. RS-2026-25617720, AI Star Fellowship (KAIST)). We also acknowledge support from the Youlchon Foundation.
% The authors at Mila acknowledge funding from the National Research Council (NRC), Samsung, and CIFAR. The research was enabled in part by computational resources provided by the Digital Research Alliance of Canada (\url{https://alliancecan.ca}) and Mila (\url{https://mila.quebec}). 
Minsu Kim acknowledges funding from the KAIST Jang Young Sil Fellow Program. We are especially grateful to Gyubin Lee and Seungju Back for their non-trivial contributions, and we thank the members of the MLML for valuable discussions and feedback throughout this project.

\section*{Broader Impacts}

GRAM studies probabilistic recursive reasoning for structured reasoning and generation. By maintaining multiple latent trajectories, it may benefit tasks such as constraint satisfaction, and scientific problem solving, where uncertainty and multiple valid solutions are common. It also suggests a way to improve reasoning through inference-time computation rather than parameter scaling alone. Its generality also entails risks: plausible but invalid generations may be mistaken for verified solutions in downstream decision-making pipelines, and multi-sample inference may increase computational and energy costs at scale. Since our experiments focus on controlled benchmarks, deployment in real-world or high-stakes settings would require rigorous validation, uncertainty calibration, and domain-specific safeguards.

\bibliographystyle{unsrtnat}
\bibliography{ref_jy}

%%%%%%%%%%%%%%%%%%%%%%%%%%%%%%%%%%%%%%%%%%%%%%%%%%%%%%%%%%%%

\newpage
\appendix

\section{Additional Method Details}

\subsection{Algorithms}
\label{appx:algorithms}

\cref{alg:training} and \cref{alg:inference} summarize training and inference. Both invoke the same latent transition, differing only in the distribution from which the stochastic guidance $\epsilon$ is drawn: the target-conditioned posterior $q_\phi$ during training, and the prior $p_\theta$ at inference, where $y$ is unavailable.

\begin{algorithm}[h]
\caption{GRAM training}
\label{alg:training}
\begin{algorithmic}[1]
\Require input $x$, target $y$;
\Require supervision steps $N_{\mathrm{sup}}$, transition steps $T$, refinement steps $K$; KL weight $\beta$; halt head $q_\psi$; 
\Procedure{Transition}{$z,\ e_x,\ \pi$}
    \State \textit{\color{gray}// $\pi$: posterior $q_\phi$ in training, prior $p_\theta$ at inference}
    \State $(h, l) \gets z$
    \For{$k = 1$ \textbf{to} $K$} \Comment{low-level refinement}
        \State $l \gets f_{\mathrm{L}}(l + h + e_x)$
    \EndFor
    \State $u \gets f_{\mathrm{H}}(h + l)$ \Comment{deterministic proposal, \cref{eq:deterministic_update}}
    \State $\epsilon \sim \pi(\cdot \mid u)$ \Comment{stochastic guidance}
    \State \Return $\big((u + \epsilon,\ l),\ u\big)$
\EndProcedure
\Statex \textit{\color{gray}// Main loop}
\State $e_x \gets f_{\mathrm{enc}}(x)$;\quad $z \gets z_0$
\For{$n = 1$ \textbf{to} $N_{\mathrm{sup}}$} 
    \State \textit{\color{gray}// T transition steps (gradient flow only at T-th step)}
    \For{$t = 1$ \textbf{to} $T-1$}
        \State $(z, \cdot) \gets \mathrm{stopgrad}\big(\textsc{Transition}(z, e_x,\ q_\phi(\cdot \mid \cdot,\, y))\big)$
    \EndFor
    \State $(z, u_T) \gets \textsc{Transition}(z, e_x,\ q_\phi(\cdot \mid \cdot,\, y))$ 
    \Statex
    \State \textit{\color{gray}// Compute loss (prediction + KL + ACT) and update weights }
    \State $(h, \cdot) \gets z$;\quad $\hat{y} \gets \argmax f_{\mathrm{dec}}(h)$;\quad $q \gets q_\psi(h)$
    \State $\mathcal{L}^{(n)} \gets \mathrm{CE}(f_{\mathrm{dec}}(h),\, y) \;+\; \beta\,\mathrm{KL}\big(q_\phi(\cdot \mid u_T, y)\ \|\ p_\theta(\cdot \mid u_T)\big)$ \Comment{only $t=T$ contributes}
    \State $\mathcal{L}^{(n)} \gets \mathcal{L}^{(n)} + \mathrm{BCE}\big(q,\ \mathbf{1}[\hat{y} = y]\big)$ \Comment{ACT halting, Appendix~\ref{appx:act}}
    \State $(\theta, \phi, \psi) \gets \textsc{AdamW}\big((\theta, \phi, \psi),\ \nabla \mathcal{L}^{(n)}\big)$
    \Statex 
    \State \textit{\color{gray}// detaching and carrying states }
    \State $z \gets \mathrm{stopgrad}(z)$ \Comment{no gradient across supervision steps}
    \State \textbf{if} $\sigma(q) > 0.5$ \textbf{then break} \Comment{halt when the prediction is judged correct}
\EndFor
\end{algorithmic}
\end{algorithm}

\begin{algorithm}[h]
\caption{GRAM inference with parallel trajectory sampling}
\label{alg:inference}
\begin{algorithmic}[1]
\Require input $x$;
\Require trajectories $N$, supervision steps $N_{\mathrm{sup}}$, transition steps $T$; \textsc{Transition} as in \cref{alg:training}
\Statex \textit{\color{gray}// Width: $N$ trajectories drawn from the prior, computed in parallel}
\State $e_x \gets f_{\mathrm{enc}}(x)$
\For{$i = 1$ \textbf{to} $N$}
    \State $z \gets z_0$
    \State \textit{\color{gray}// Depth: fixed number of steps (halting disabled for fair comparison)}
    \For{$n = 1$ \textbf{to} $N_{\mathrm{sup}}$}
        \For{$t = 1$ \textbf{to} $T$}
            \State $(z, \cdot) \gets \textsc{Transition}(z, e_x,\ p_\theta(\cdot \mid \cdot))$ \Comment{prior: $y$ is unavailable}
        \EndFor
    \EndFor
    \State $(h, \cdot) \gets z$;\quad $\hat{y}^{(i)} \gets \argmax f_{\mathrm{dec}}(h)$
\EndFor
\Statex
\State \textit{\color{gray}// Select final prediction among $N$ candidates}
\State \Return $\argmax_{y} \sum_{i=1}^{N} \mathbf{1}[\hat{y}^{(i)} = y]$ \Comment{majority voting}
\end{algorithmic}
\end{algorithm}

\subsection{Empirical Validation of the Surrogate Objective}
\label{appx:approx_analysis}

We further analyze the approximation introduced by the surrogate training objective $\mathcal{L}_{\mathrm{GRAM}}$ used in \cref{sec:training}, both qualitatively and empirically.

\textbf{Truncation as a gradient approximation.}
We frame $\mathcal{L}_{\mathrm{GRAM}}$ as a gradient approximation rather than a separate variational objective. The full trajectory-level ELBO (\cref{eq:traj_kl_eps}) involves a sum of KL terms across all $T_{\mathrm{Total}}$ transitions, and computing its exact gradient requires backpropagation through the entire trajectory. To enable training with constant memory, we propagate gradients only through the final transition of each supervision step. This is a standard practice in recurrent latent variable models with long computation chains: ELBOs over truncated sequences are used, for example, in VRNN~\citep{vrnn} and SRNN~\citep{srnn}, while truncated latent imagination is used in Dreamer-family world models~\citep{dreamerv2, dreamerv3}. Trading a small gradient bias for training stability via local truncation is therefore well-precedented; what is specific to GRAM is applying this approximation at the level of recursive reasoning trajectories rather than temporal sequences.

\begin{figure}[h]
    \centering
    \includegraphics[width=0.9\linewidth]{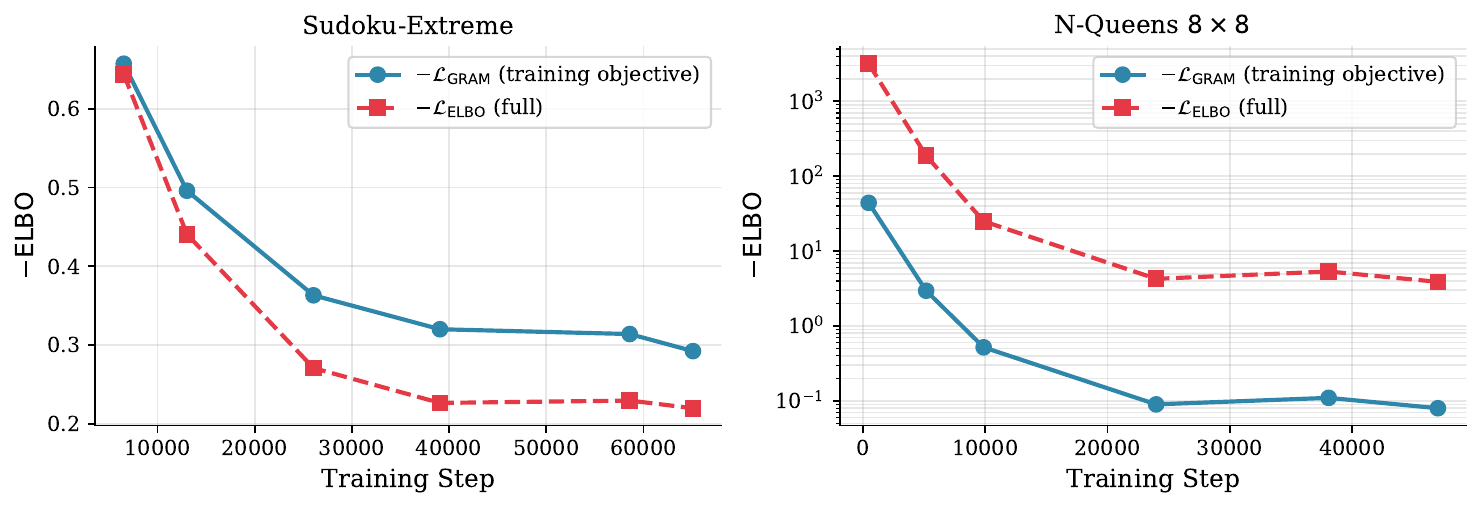}
    \caption{\textbf{Full ELBO $\mathcal{L}_{\mathrm{ELBO}}$ and surrogate objective $\mathcal{L}_{\mathrm{GRAM}}$ throughout training} (plotted as $-\mathrm{ELBO}$, smaller is better). On both Sudoku-Extreme (left) and N-Queens $8\times 8$ (right), both quantities decrease monotonically over training, indicating that gradient updates of $\mathcal{L}_{\mathrm{GRAM}}$ consistently improve the full variational bound. The two curves do not coincide because $\mathcal{L}_{\mathrm{ELBO}}$ sums KL contributions across all $T_{\mathrm{Total}}$ transitions while $\mathcal{L}_{\mathrm{GRAM}}$ evaluates only the final-step KL of each supervision step; their gap reflects the cumulative KL across earlier transitions, not a failure of optimization. The N-Queens plot uses a log scale on the y-axis due to the large dynamic range.}
    \label{fig:elbo_comparison}
\end{figure}

\textbf{Empirical validation.}
To verify that optimizing $\mathcal{L}_{\mathrm{GRAM}}$ effectively drives improvement in the full variational bound, we compute both quantities on the validation set throughout training. The full ELBO $\mathcal{L}_{\mathrm{ELBO}}$ is evaluated as in \cref{eq:traj_kl_eps}, summing the reconstruction term and KL contributions across all $T_{\mathrm{Total}}$ transitions; the surrogate objective is evaluated as the average of $\mathcal{L}_{\mathrm{GRAM}}^{(n)}$ over the $N_{\mathrm{sup}}$ supervision steps. \cref{fig:elbo_comparison} reports the results on Sudoku-Extreme and N-Queens $8\times 8$.

Both $\mathcal{L}_{\mathrm{ELBO}}$ and $\mathcal{L}_{\mathrm{GRAM}}$ improve monotonically throughout training on both tasks. This indicates that, despite the truncation, gradient updates of $\mathcal{L}_{\mathrm{GRAM}}$ effectively drive improvement in the full variational bound. Since $\mathcal{L}_{\mathrm{ELBO}}$ also serves as an indirect estimate of the negative log-likelihood, its consistent improvement provides evidence that GRAM optimizes a well-defined data likelihood, even though training relies on the surrogate.

The gap between the two curves in \cref{fig:elbo_comparison} reflects the structural difference between the two quantities — $\mathcal{L}_{\mathrm{ELBO}}$ accumulates KL terms across all transitions while $\mathcal{L}_{\mathrm{GRAM}}$ evaluates only the final-step KL of each supervision step — rather than an optimization failure. This gap is consistent with $\mathcal{L}_{\mathrm{GRAM}}$ being a biased but useful surrogate for $\mathcal{L}_{\mathrm{ELBO}}$.

\subsection{Adaptive Computation Time}
\label{appx:act}
 
GRAM optionally adopts adaptive computation time (ACT)~\citep{hrm,trm,universal} at inference, allowing each trajectory to terminate at a learned halting depth rather than running for a fixed number of supervision steps. We follow the Q-learning formulation introduced by HRM~\citep{hrm} and adopted in TRM~\citep{trm}.
 
\textbf{Halt head.}
The decoder includes an auxiliary head $q_\psi : \mathbb{R}^{D} \to \mathbb{R}^{2}$ that maps the high-level state $h$ to two scalar values, $q_\psi(h) = (q^{\mathrm{halt}}, q^{\mathrm{continue}})$. These are interpreted as estimated Q-values for the binary action of halting or continuing computation at the current supervision step.
 
\textbf{Training.}
The halt head is trained jointly with the main objective via a temporal-difference loss. After computing the latent state $z_T^{(n)}$ at the end of supervision step $n$, we form Q-learning targets:
\begin{itemize}
    \item $\hat{q}^{\mathrm{halt}}_n = \mathbf{1}[\hat{y}^{(n)} = y]$, indicating whether decoding the current state would yield a correct prediction.
    \item $\smash{\hat{q}^{\mathrm{continue}}_n = \max\!\left(q^{\mathrm{halt}}_{n+1},\, q^{\mathrm{continue}}_{n+1}\right)}$, the bootstrapped value of running one more supervision step.
\end{itemize}
The halt head is trained by regression to these targets:
\begin{align}
    \mathcal{L}_{\mathrm{ACT}}
    =
    \sum_{n=1}^{N_{\mathrm{sup}}}
    \Big[
        \big(q^{\mathrm{halt}}_n - \hat{q}^{\mathrm{halt}}_n\big)^2
        +
        \big(q^{\mathrm{continue}}_n - \hat{q}^{\mathrm{continue}}_n\big)^2
    \Big].
\end{align}
This auxiliary loss is added to the main training objective.
 
\textbf{Inference.}
At inference, computation proceeds one supervision step at a time. After each step $n$, we evaluate $q_\psi(h^{(n)})$ and halt if $q^{\mathrm{halt}}_n > q^{\mathrm{continue}}_n$, returning $\hat{y}^{(n)}$ as the prediction. Otherwise, computation continues to the next supervision step, up to a maximum budget of $N_{\mathrm{sup}}^{\max}$ steps. Different trajectories sampled in parallel may therefore terminate at different depths, complementing the parallel-sampling scheme described in \cref{sec:inference}. In practice, we found that using only $q^{\mathrm{halt}}$ (halting when $\sigma(q^{\mathrm{halt}}) > 0.5$, without the continue branch) performs comparably while simplifying implementation; our released code uses this variant.

\subsection{Latent Process Reward Model (LPRM)}
\label{appx:lprm}

Parallel sampling produces $N$ candidate outputs, and any rule that scores candidates can be used to select among them. We use majority voting throughout this paper. However, we also discuss a learned selector supported by our implementation, the Latent Process Reward Model (LPRM).

The LPRM is a value head $v_\psi(z_t)$ trained to predict the expected accuracy of the final output, conditioned on the current latent state $z_t$. It is trained jointly with the main objective via a regression loss
\begin{align}
\mathcal{L}_{\mathrm{LPRM}} = \sum_{t=1}^{T} \left(v_\psi(z_t) - r\right)^2,
\end{align}
where $r \in \{0, 1\}$ denotes the accuracy of the final prediction for a given trajectory. At inference, best-of-$N$ selection returns $\hat{y}^{(n^*)}$ with $n^* = \argmax_n v_\psi(z_T^{(n)})$, in place of the majority-voted prediction.

\section{Training and Architecture Details}
\subsection{Architecture Details}
\label{appx:arch_detail}
GRAM consists of three components: Encoder, Recursive Core, and Decoder.

\begin{table}[h]
\centering
\caption{Architecture components.}
\label{tab:arch_components}
\small
\begin{tabular}{@{}lll@{}}
\toprule
\textbf{Component} & \textbf{Module} & \textbf{Description} \\
\midrule
\multicolumn{3}{l}{\textit{Encoder}} \\
\quad & Token Embedding & vocab $\rightarrow D$ \\
\quad & Puzzle Embedding & 16 tokens  \\
\quad & Position Encoding & RoPE \\
\midrule
\multicolumn{3}{l}{\textit{Recursive Core}} \\
\quad & $f_L$, $f_H$ & [Attn. + SwiGLU] $\times$ 2 blocks \\
\quad & Iterations & $K$ low-level, $T$ high-level steps \\
\quad & Prior head $\mu_\theta, \sigma_\theta$ & SwiGLU MLP, input $u_t$ \\
\quad & Posterior head $\mu_\phi, \sigma_\phi$ & SwiGLU MLP, input $[\,u_t \,;\, \mathrm{repr}(y)\,]$ \\
\midrule
\multicolumn{3}{l}{\textit{Decoder}} \\
\quad & LM Head & SwiGLU MLP($D \rightarrow$ vocab) \\
\quad & Q Head & Linear($D \rightarrow 2$) for halt \\
\bottomrule
\end{tabular}
\end{table}
\textbf{Encoder.} Input tokens are mapped to embeddings via a token embedding layer and combined with positional encodings (RoPE)~\citep{rope}. The embeddings are scaled by $\sqrt{D}$ and prepended with 16 puzzle embedding tokens~\citep{hrm}, produced from a learned embedding indexed by a puzzle identifier. On ARC-AGI~\citep{arc, arc2}, each puzzle has its own identifier, so these tokens distinguish tasks. On all other tasks, a single identifier is shared across examples, and they instead act as CLS-like register tokens. These tokens are carried through the recursion, so the latent states $h$ and $l$ span these positions as well as the input tokens.

\textbf{Recursive Core.} The core maintains two latent states: $h$ (high-level) and $l$ (low-level). For each outer step, the low-level state is refined $K$ times via $l \leftarrow f_L(l + h + e_x)$, injecting the input embedding at each iteration. The high-level state is then updated via $h \leftarrow f_H(h + l)$. $f_L$ and $f_H$ are implemented by a single network whose parameters are reused for both roles and across all latent transitions and supervision steps, so recursion depth can be increased without increasing the parameter count. In addition, as an exception, we use [SwiGLU + SwiGLU] network for the Recursive Core module instead of [Attention + SwiGLU] for Sudoku tasks, following~\citep{trm}. For initialization of $z_0=(h_0, l_0)$, we sample once from the standard Gaussian distribution $\mathcal{N}(0, I)$, then save the value within the network checkpoint and load it again, meaning the initialized $z_0$ has a fixed value.

\textbf{Stochastic Guidance Heads.} The prior heads $(\mu_\theta, \sigma_\theta)$ and the posterior heads $(\mu_\phi, \sigma_\phi)$ are implemented as token-wise SwiGLU~\citep{swiglu} MLPs, applied independently at each of the $L$ sequence positions of the high-level state. The two have identical architecture but separate weights, and differ in their input: the prior head takes the deterministic update $u_t$, while the posterior head takes $u_t$ concatenated along the feature dimension with a representation of the target $y$. For binarized-MNIST~\citep{mnist}, this representation is produced by the convolutional patch encoder described in \cref{tab:mnist_arch}; for all other tasks, it is a one-hot encoding of $y$ over the task vocabulary (\cref{tab:task_config}).

\textbf{Decoder.} The decoder extracts content tokens from $h$ (excluding puzzle embedding positions) and maps them to logits via a SwiGLU MLP head. An auxiliary head predicts halt decisions from the first puzzle embedding position of $h$.

\paragraph{Encoder and Decoder for Image Patches.} In the MNIST~\citep{mnist} image generation task, we first construct a binarized dataset by normalizing the original discrete pixel values ($0 \sim 255$) to the continuous range $[0, 1]$ and applying a threshold at $0.5$. For the network architecture, we employ a convolutional patch encoder, following~\citep{d3pm_pytorch, dit}.

The encoding process proceeds in three stages. First, the discrete input tokens $x \in \{0, 1\}$ are normalized to the range $[-1, 1]$. Second, to capture local spatial dependencies before patchification, the normalized image passes through a shallow convolutional encoder. This encoder consists of two stacked blocks, where each block comprises a 2D convolution~\citep{cnn, alexnet} with a $5 \times 5$ kernel and padding 2, a SiLU non-linearity~\citep{silu}, and Group Normalization (GN)~\citep{group_norm}. Finally, the resulting feature map is divided into non-overlapping patches of size $P \times P$ and linearly projected to match the model's hidden dimension $D$. The detailed architectural specifications and dimension transitions are summarized in Table~\ref{tab:mnist_arch}.

\begin{table}[h]
\centering
%\small % Make text smaller and compact
\caption{Detailed architecture of the Image Patch Encoder for binarized-MNIST. $H, W$ denote image resolution, $C$ input channels, $P$ patch size, $N_p$ the number of patches, and $D$ the hidden dimension.}
\label{tab:mnist_arch}
\vspace{0.1cm}
\begin{tabular}{l|l|c}
\toprule
\textbf{Stage} & \textbf{Layer / Operation} & \textbf{Output Dim.} \\
\midrule
\multirow{2}{*}{1. Norm.} & Input Tokens & $(B, C, H, W)$ \\
 & Linear Scaling $[-1, 1]$ & $(B, C, H, W)$ \\
\midrule
\multirow{4}{*}{2. Conv} & Conv2d $5\times5$ ($p=2$) & \multirow{2}{*}{$(B, D/2, H, W)$} \\
 & SiLU $\rightarrow$ GN(32) & \\
 \cmidrule(l){2-3}
 & Conv2d $5\times5$ ($p=2$) & \multirow{2}{*}{$(B, D/2, H, W)$} \\
 & SiLU $\rightarrow$ GN(32) & \\
\midrule
\multirow{2}{*}{3. Patch} & Flatten Patches & $(B, N_{p}, P^2 \cdot \frac{D}{2})$ \\
 & Linear Projection & $(B, N_{p}, D)$ \\
\bottomrule
\end{tabular}
\end{table}

\textbf{Hyperparameters.} Following~\citep{hrm, trm}, both the input and output are represented as sequences of shape $[B, L]$,
where $B$ denotes the batch size and $L$ the context length.  Each input sequence includes 16 fixed puzzle embedding tokens.
The latent states $h_t$ and $l_t$, as well as the decoder output, have shape $[B, L, D]$,
with embedding dimension $D$. The Transformer~\citep{transformer} backbone uses embedding dimension $D=512$, attention heads $N_{\text{head}}$=8 , and FFN hidden dimension $D_h$=512. Each latent transition $z_t \to z_{t+1}$ applies $K$ low-level (inner) steps followed by one high-level update, with $K=6$ for Sudoku-Extreme~\citep{hrm} and $K=4$ for all other tasks. Each supervision step consists of $T=3$ such transitions.

\subsection{Training Details}
\label{appx:train_detail}

\textbf{Task Configuration.}
All tasks represent inputs and outputs as discrete token sequences (Summarized in~\cref{tab:task_config}). Puzzle identifiers are task-specific only for ARC-AGI, and shared across examples otherwise.
\begin{itemize}
    \item For \textit{Sudoku-Extreme}~\citep{hrm}, the 9$\times$9 grid is flattened row-by-row into 81 tokens with vocabulary size 11 (0=pad, 1=blank, 2--10=digits). Unconditional Sudoku generation uses the same encoding, with an all-blank board as input and a complete solution as target.
    \item For \textit{ARC-AGI}~\citep{arc,arc2}, variable-size grids are padded to a fixed 30$\times$30 canvas with EOS markers, yielding 900 tokens and vocabulary size 12 (0=pad, 1=eos, 2--11=colors); task-specific puzzle embeddings are prepended to distinguish different ARC tasks. 
    \item \textit{N-Queens} flattens an $N \times N$ board row-by-row into $N^2$ tokens with vocabulary size 3 (0=pad, 1=empty, 2=queen). 
    \item \textit{Graph Coloring} encodes the upper triangle of the adjacency matrix as $\nicefrac{n(n+1)}{2}$ tokens, using 0=PAD, 1=no-edge, and 2=edge for inputs and $3 + \text{color\_id}$ for output colors.
    \item For image generation on \textit{binarized-MNIST}~\citep{mnist}, images are quantized into pixels with vocabulary size 3 (0=pad, 1=black, 2=white) and processed via CNN-based patchification~\citep{cnn, dit}, with the encoder applying patchify and the decoder unpatchify. Then, patched input forms $14\times 14$ flattened sequence tokens.
\end{itemize}

\begin{table}[h]
\centering
\caption{Task-specific configurations.}
\label{tab:task_config}
\small
\begin{tabular}{@{}lcccc@{}}
\toprule
\textbf{Dataset} & \textbf{Seq. Len} & \textbf{Vocab} & \textbf{Encoding} \\
\midrule
Sudoku & 81 & 11 & 9$\times$9 grid, row-major \\
ARC-AGI & 900 & 12 & 30$\times$30 padded canvas \\
N-Queens & $N^2$ & 3 & $N \times N$ board \\
Graph Coloring & $\frac{n(n+1)}{2}$ & 6 & Adjacency upper triangle \\
binarized-MNIST & 196 & 3 & 14$\times$14 patches \\
\bottomrule
\end{tabular}
\end{table}

\textbf{Training Details.}
We train GRAM using AdamW~\citep{adamw} with a learning rate of $10^{-4}$ and a weight decay of 1.0. The global batch size is 768. For stability, we apply exponential moving average (EMA) with decay 0.999, following~\citet{brock2018large} and \citet{song2020improved}. To prevent posterior collapse, we use a KL balance~\citep{dreamerv2, dreamerv3} coefficient of 0.8. The number of deep supervision steps is $N_{\text{sup}}=16$ for all tasks. The KL coefficient $\beta$ is set to $0.1$ (Sudoku), $0.04/0.1$ (ARC-AGI-1/2), $0.07/0.045$ (N-Queens $8\times8$/$10\times10$), $0.5/0.45$ (Graph Coloring with 8/10 nodes), and $0.015$ (binarized-MNIST). Task-specific training configurations are summarized in \cref{tab:training_config}.

\textbf{binarized-MNIST Model and Training Configurations.}
For the comparison, we select VAE, TRM, D3PM, and GRAM configurations with comparable parameter counts. The VAE uses a 2-layer convolutional encoder with a base channel width of 128, separate MLP heads for the posterior mean and log-variance, a 384-dimensional latent space, and a transposed-convolutional decoder, resulting in 15.81M parameters. D3PM uses a convolutional patch encoder followed by a 3-layer DiT-style Transformer with hidden dimension 512 and 8 attention heads, where diffusion timesteps condition each block through adaptive layer normalization (adaLN)~\citep{dit}. It uses an absorbing-state corruption process with 1,000 diffusion steps and contains 18.07M parameters. GRAM uses a 2-layer recursive Transformer core with hidden dimension 512, 8 attention heads, and a 256-dimensional stochastic latent space, resulting in 15.82M parameters. TRM, the deterministic recursive baseline, has 10.90M parameters.

Because D3PM and GRAM treat pixels as categorical variables, we train the VAE with a binary cross-entropy reconstruction term to maintain a consistent binary-pixel representation across the three models.

\textbf{Unconditional Sudoku Generation Model and Training Configurations.}
Unlike the conditional Sudoku-solving setting, where the input board contains given clues, the model receives an entirely blank board and samples a complete $9 \times 9$ Sudoku board from its learned prior. We evaluate each generated board using the standard Sudoku validity criterion: every row, column, and $3 \times 3$ box must contain the digits 1 through 9 exactly once. We report the validity rate over 100K generated boards. To check whether high validity comes from repeatedly producing the same board, we also compute the fraction of unique boards among valid samples.

For GRAM, we construct the unconditional training set from Sudoku-Extreme~\citep{hrm}, the Sudoku benchmark used by HRM and TRM. We sample 50K complete solutions from the original training split, discard the clue patterns, and use an all-blank board as input with the complete solution as the target. No data augmentation is used. We train GRAM on this derived 50K-solution set for 200 epochs with learning rate $10^{-4}$, EMA decay 0.999, and KL coefficient 0.05. The resulting model contains 10.9M parameters and uses 16 inference steps.

For D3PM, we use a DiT-style Transformer backbone with the absorbing-state noise process, and evaluate two model sizes. D3PM-Big uses hidden dimension 768, 5 Transformer blocks, and 12 attention heads, yielding 55.1M parameters, while D3PM-Small uses hidden dimension 512, 3 Transformer blocks, and 8 attention heads, yielding 15.9M parameters. Both variants are trained on the same derived training set and generate boards with 1000 denoising steps.

\begin{table}[h]
\centering
\caption{Training configurations on NVIDIA RTX 4090 GPUs.}
\label{tab:training_config}
\small
\begin{tabular}{@{}lccc@{}}
\toprule
\textbf{Task} & \textbf{Epochs} & \textbf{GPUs} & \textbf{Time} \\
\midrule
Sudoku-Extreme & 50K & 8 & 2h \\
ARC-AGI & 200K & 8 & 5 days \\
N-Queens (8$\times$8) & 3K & 8 & 1h \\
N-Queens (10$\times$10) & 1K & 8 & 3h \\
Graph Coloring (8 nodes) & 5K & 8 & 1.5h \\
Graph Coloring (10 nodes) & 5K & 8 & 6h \\
Binarized-MNIST & 1.8K & 8 & 16h \\
Sudoku-Generation & 0.2K & 4 & 0.6h \\
\bottomrule
\end{tabular}
\end{table}

\section{Additional Details of Experiment Setup }
\label{sec:experiment_details}

\subsection{Challenging Puzzle Tasks}

\paragraph{ARC-AGI Evaluation Protocol.}
\label{appx:arc_protocol}

We follow the standard ARC-AGI evaluation protocol used by HRM~\citep{hrm} and TRM~\citep{trm}. Each test input is transformed into 1,000 variants by combining rotations and reflections of the grid with permutations of its colors. The model is run on each variant, each prediction is mapped back by the inverse transformation, and the predictions are aggregated by voting; accuracy is reported as pass@2, the official ARC-AGI metric, under which a task counts as solved if either of the two submitted attempts is correct. For GRAM, we draw a single sample per augmented input, so the ARC-AGI numbers in \cref{tab:puzzle} use neither parallel sampling nor trajectory selection. Augmentation is used only for ARC-AGI; all other benchmarks are evaluated on the original inputs. Appendix~\ref{app:scaling} isolates the effect of parallel sampling on ARC-AGI by removing augmentation.

\paragraph{Looped TF on ARC-AGI.}
\label{appx:looped_tf_arc}
We report Looped Transformer~\citep{looped_transformer} results on Sudoku-Extreme but omit them on ARC-AGI due to prohibitive training cost. Under the same setup used for the other recursive models on Sudoku-Extreme (batch size 768, on $8\times$ NVIDIA RTX 4090 GPUs), training Looped TF on Sudoku-Extreme already takes 19 hours. Measuring its per-step time directly on ARC-AGI and scaling to the 200K-epoch budget used for GRAM on ARC-AGI gives approximately \textbf{97 days} ($\approx 776$ GPU-days) for a full training run.
 
This gap stems from two compounding factors. First, Looped TF lacks deep supervision: HRM, TRM, and GRAM perform $N_{\mathrm{sup}}$ gradient updates per trajectory (one per segment), whereas Looped TF performs only one update at the end of the full trajectory, slowing convergence. Second, Looped TF lacks adaptive halting such as ACT~\citep{hrm,trm,universal}, so every input must be processed for the maximum recursion depth, increasing per-example sequential compute. Both inefficiencies compound at ARC-AGI scale, making a full Looped TF training run impractical.

\subsection{Multi-solution Puzzle Tasks}
\label{sec:multi_solution_details}

\paragraph{N-Queens Problem.}

\begin{figure}[h]
\centering
\includegraphics[width=0.8\linewidth]{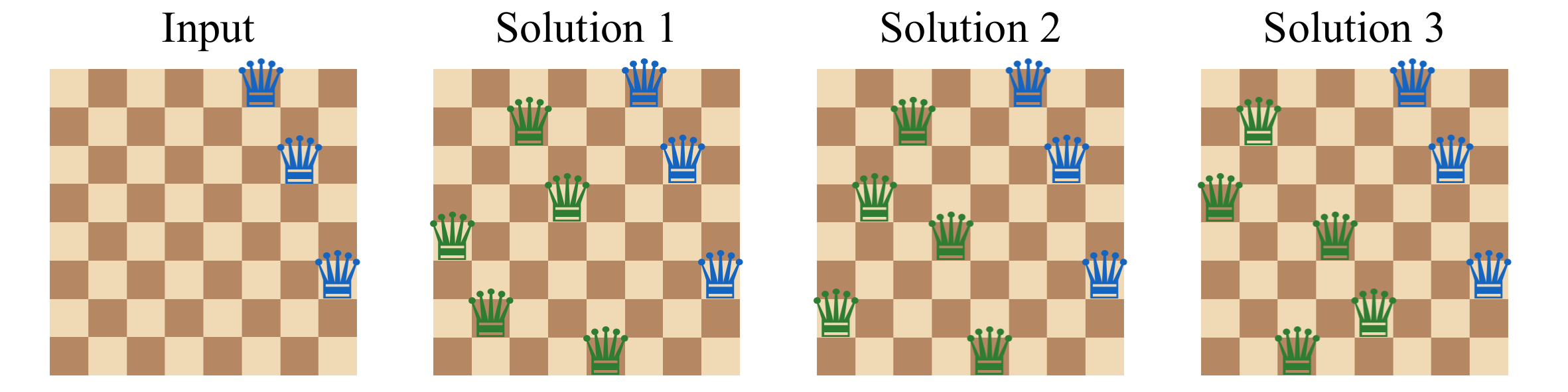}
\caption{\textbf{Example of an $8\times8$ N-Queens puzzle instance.} In this example, 5 queens are removed from the full board, leaving 3 queens. The model must find the positions of the remaining queens. This configuration admits exactly 3 valid solutions.}
\label{fig:nqueens_example}
\end{figure}
The N-Queens problem~\citep{nqueens} requires placing $N$ queens on an $N \times N$ chessboard such that no two queens attack each other, meaning no queens share the same row, column, or diagonal. Figure~\ref{fig:nqueens_example} illustrates an example where 5 queens are removed from an $8 \times 8$ solution, resulting in a puzzle with 3 distinct valid completions.

\begin{figure}[ht]
\centering
\includegraphics[width=0.8\linewidth]{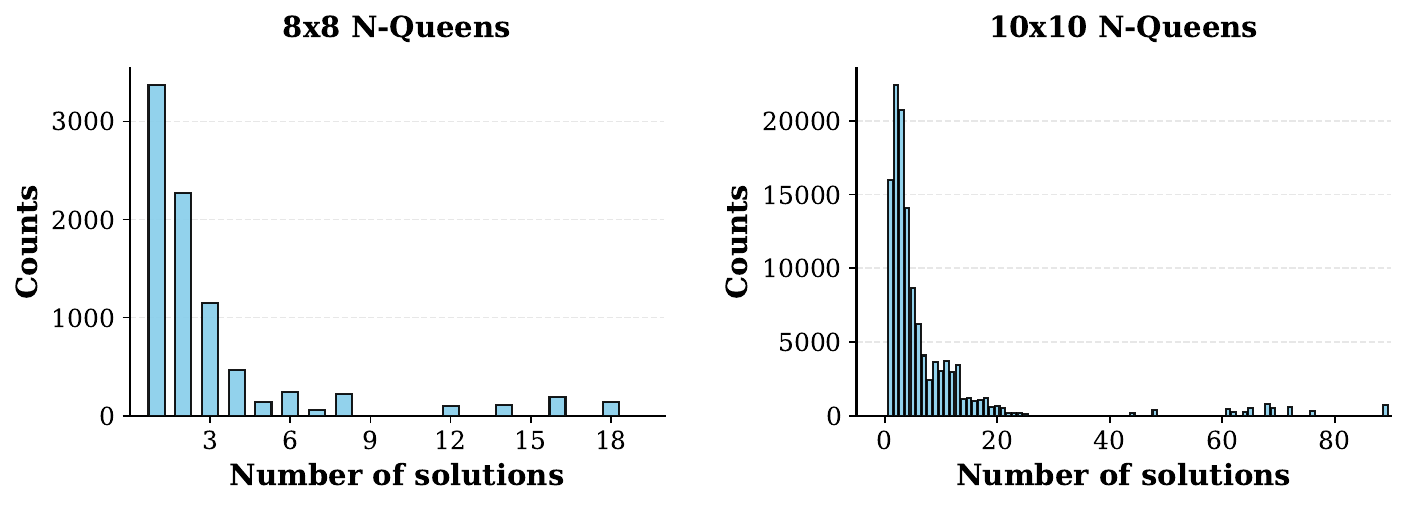}
\caption{\textbf{Distribution of the number of valid solutions for generated N-Queens instances.} The dataset covers a wide range of solution counts, testing the model's ability to recover multiple valid outputs.}
\label{fig:nqueens_solution_dist}
\end{figure}

To construct the dataset, we first generated all valid complete N-Queens solutions for $N=8$ and $N=10$ by backtracking, yielding 92 and 724 solutions respectively. We then created puzzle instances by removing every combination of a fixed number of queens, treating the remaining partial configuration as the input and the original complete board as the target label. To generate instances yielding diverse valid completions, we removed $k \in \{5, 6, 7\}$ queens for the $8 \times 8$ setting and $k \in \{7, 8, 9\}$ queens for the $10 \times 10$ setting. Since a partial configuration can arise from several solutions, an input may carry multiple targets, on average 1.65 for $8 \times 8$ and 2.93 for $10 \times 10$. The distribution of solution counts for our generated dataset is shown in Figure~\ref{fig:nqueens_solution_dist}.

For evaluation, we employed an 85:15 train-test split. Crucially, to prevent data leakage and ensure the model learns to reason rather than memorize, the split was performed based on unique \emph{input} configurations. This guarantees that no input pattern in the test set appears in the training set. The resulting dataset contains 7,224 training and 1,240 test instances for $8 \times 8$, and 108,234 and 18,466 for $10 \times 10$. Inputs are flattened into discrete 1D sequences $x \in \{0, 1, 2\}^L$, where $L=N^2$, along with zero-padded puzzle embedding tokens. Vocabulary mapping follows: padding (0), empty (1), and queen (2). We use neither data augmentation nor rebalancing between single- and multi-solution instances.

\paragraph{Graph Coloring Problem.}

\begin{figure}[ht]
\centering
\includegraphics[width=0.8\linewidth]{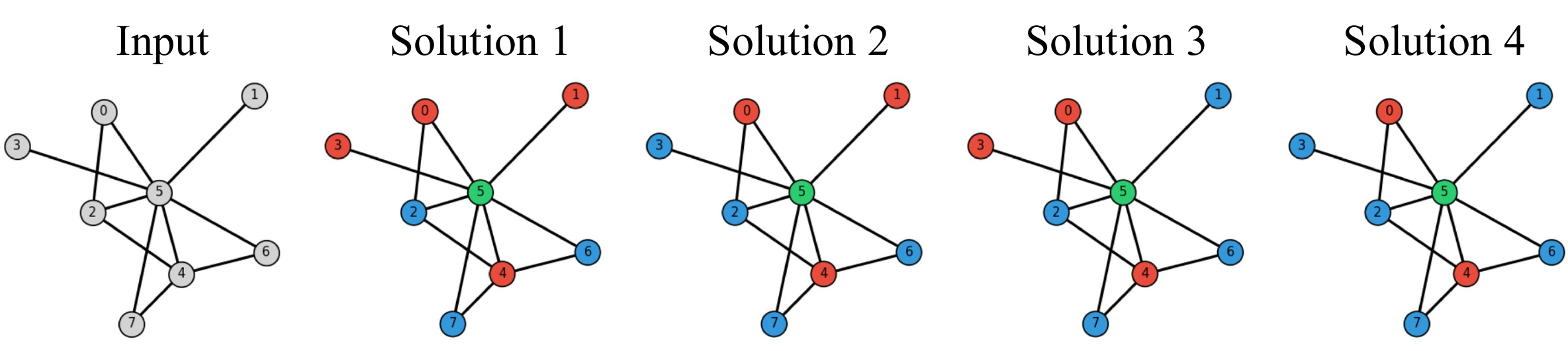}
\caption{\textbf{Graph Coloring Example}}
\label{fig:graph_coloring_example}
\end{figure}

The Graph Coloring problem requires assigning one of $k$ colors to each node in a graph such that no two adjacent nodes share the same color. We consider graphs with $N \in \{8, 10\}$ nodes and use $k=3$ colors. Figure~\ref{fig:graph_coloring_example} illustrates an example instance with $N=8$ nodes and $k=3$ colors.

Graphs are generated using the Erdős–Rényi random graph model~\citep{erdos}, following the generation pipeline from GNN-GCP~\citep{gnngcp}, in which a CP-SAT solver verifies the chromatic number of each sampled graph. We retain only graphs whose chromatic number is exactly 3. The resulting graphs have 6 to 14 edges for $N=8$ and 9 to 19 edges for $N=10$. Mean edge density is 0.38 in training and 0.42 in test for $N=8$, and 0.28 and 0.34 for $N=10$.

\begin{figure}[ht]
\centering
\includegraphics[width=0.8\linewidth]{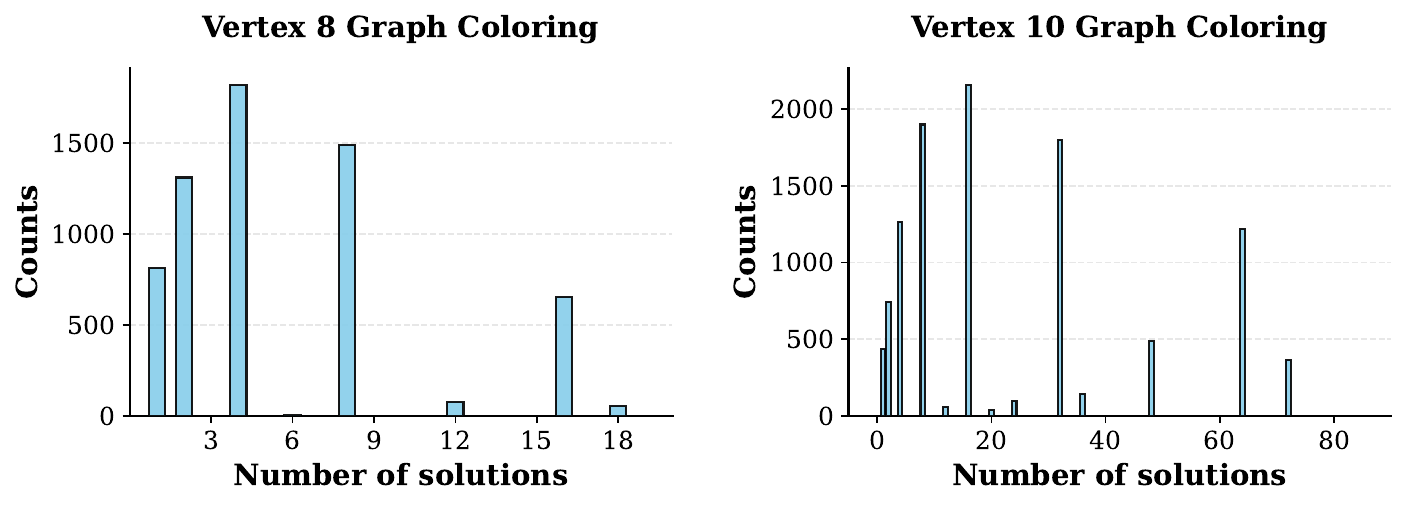}
\caption{\textbf{Distribution of the number of valid solutions for generated graph coloring instances.} The dataset covers a wide range of solution counts, testing the model's ability to recover multiple valid outputs.}
\label{fig:graph_solution_dist}
\end{figure}

For each graph, we enumerate all valid 3-colorings and retain only canonical forms, in which colors are relabelled in order of first appearance, to eliminate redundant solutions under color permutation. Since each graph admits several canonical colorings, a training graph may carry multiple targets, on average 3.40 for $N=8$ and 8.73 for $N=10$; the test split uses one coloring per graph. The distribution of solution counts is shown in Figure~\ref{fig:graph_solution_dist}.

The final dataset consists of 7,002 training and 255 test instances for $N=8$, and 13,465 training and 192 test instances for $N=10$. The dataset is publicly available at \url{https://huggingface.co/datasets/CUN-bjy/graph-coloring}.

\textbf{Input and Output Representation.} The input graph is represented by extracting the upper triangular portion of the adjacency matrix (including the diagonal) and flattening it into a 1D sequence of $N(N+1)/2$ tokens. The output is a sequence of length $N$, where each position encodes the assigned color for the corresponding node, followed by PAD tokens up to the input length. Vocabulary mapping is as follows: PAD (0), no edge (1), edge (2), and colors (3, 4, 5) for red, blue, and green respectively.

\section{Additional Experimental Results}

\subsection{Additional Results on Challenging Puzzle Benchmarks}
\label{app:puzzle_analysis}

\cref{tab:puzzle} reports test accuracy on three challenging puzzle benchmarks. Here we provide additional observations complementing the main text.

\paragraph{GRAM Advances the Recursive-Reasoning Line.}
Across all three benchmarks, GRAM outperforms prior recursive baselines. On Sudoku-Extreme, GRAM already surpasses TRM with a single sample (86.6\% against 83.6\%), and parallel sampling, an axis the stochastic transition makes available and deterministic recursion does not have, raises it further to 96.1\%. At matched inference compute, GRAM reaches 96.1\% against TRM's 92.1\% (Appendix~\ref{appx:flops}). The complete failure of direct prediction on Sudoku and ARC-AGI-2 (0\% in both cases) further confirms that recursive computation is essential for these tasks --- single-pass models, regardless of capacity, cannot solve them. Together, these results indicate that GRAM's gains arise from organizing recursive computation as probabilistic, multi-trajectory reasoning.

\paragraph{Sudoku-Extreme Resists Parameter Scaling.}
All tested large reasoning models (LRMs), including Deepseek-R1 (671B), score 0\% on Sudoku-Extreme. This suggests that pretrained capacity alone does not transfer to constraint-propagation reasoning, and that benchmarks like Sudoku-Extreme probe a fundamentally different axis from those captured by general-purpose LRMs. On ARC-AGI, more recent LRMs such as Gemini 3 Pro (75.0\% on ARC-1, 31.1\% on ARC-2) remain substantially ahead of all recursive models, highlighting that abstract few-shot reasoning still benefits from scale; we view these numbers as benchmark-difficulty reference points rather than controlled baselines.

\begin{table}[t]
\caption{\textbf{Test accuracy (\%) on Challenging Puzzle Benchmarks.} All recursive model scores were obtained at 16 supervision steps; ARC-AGI accuracy is pass@2 following the standard protocol (Appendix~\ref{appx:arc_protocol}). Sudoku-Extreme scores of recursive models are averaged over 3 seeds and reported as mean $\pm$ standard deviation. ARC-AGI scores of HRM and TRM, all Direct Pred scores, and the Sudoku-Extreme scores of large reasoning models are taken from the original papers~\citep{hrm, trm}, with ARC-AGI on the public evaluation set. ARC-AGI scores of large reasoning models are from the ARC Prize leaderboard~\citep{arcprize2026arcagi_benchmarking} on the semi-private set. Bold marks the best result within each group.}
\label{tab:puzzle}
\centering
\large 
\setlength{\tabcolsep}{3pt}
\renewcommand{\arraystretch}{1.08}
\setlength{\heavyrulewidth}{1pt} 

\begin{adjustbox}{max width=0.55\linewidth}
\begin{tabular}{@{}lcccc@{}}
\toprule
\textbf{Method} & \textbf{\#Params} & \textbf{Sudoku} & \textbf{ARC-1} & \textbf{ARC-2} \\
\midrule
\textit{\textcolor{gray}{Large Reasoning Models}} \\[2pt]
\quad Deepseek-R1 & 671B & 0.0 & 15.8 & 1.3 \\
\quad Claude 3.7 16k & N/A & 0.0 & 28.6 & 0.7 \\
\quad o3-mini-high & N/A & 0.0 & 34.5 & 3.0 \\
\quad GPT 5.2 (low) & N/A & -- & 55.7 & 9.7 \\
\quad Grok-4-thinking & 1.7T & -- & 66.7 & 16.0 \\
\quad Gemini 3 Pro & N/A & -- & \textbf{75.0} & \textbf{31.1} \\
\midrule
\textit{\textcolor{gray}{Recursive Models}} \\[2pt]
\quad Direct Pred & 27M & 0.0 & 21.0 & 0.0 \\
\quad Looped TF & 7M & 61.7\cspstd{0.9} & - & - \\
\quad HRM & 27M & 56.8\cspstd{1.2} & 40.3 & 5.0 \\
\quad TRM & 7M & 83.6\cspstd{1.3} & 44.6 & 7.8 \\
\midrule
\rowcolor{blue!8} \quad \textbf{GRAM (Ours)}, $N{=}1$ & 10M & \textbf{86.6}\cspstd{2.2} & \textbf{50.3}\cspstd{2.3} & \textbf{9.5}\cspstd{2.5} \\
\rowcolor{blue!8} \quad\quad $+$ parallel sampling ($N{=}20$) & & \textbf{96.1}\cspstd{0.9} & -- & -- \\
\bottomrule
\end{tabular}
\end{adjustbox}
\end{table}

\subsection{Computational Cost of Inference-Time Scaling}
\label{appx:flops}

We measure the computational cost of inference-time scaling on Sudoku-Extreme. Floating-point operations are counted per puzzle with \texttt{torch.utils.flop\_counter}, and wall-clock latency is measured on identical hardware with a single puzzle in flight, so that parallel samples are batched into one forward pass while recursion steps execute sequentially. Accuracies are the 3-seed means reported in \cref{fig:scaling}.

\paragraph{Stochastic transitions add little per-step cost.}
\cref{tab:flops_per_iter} reports inference FLOPs per puzzle. The stochastic transition introduces only the noise head, which produces the mean and variance of the latent perturbation, and leaves the recursive core unchanged. GRAM therefore costs 3.5\% more per iteration than TRM.

\begin{table}[h]
\caption{Inference FLOPs per puzzle in TFLOPs.}
\label{tab:flops_per_iter}
\centering
\small
\begin{tabular}{@{}lcc@{}}
\toprule
\textbf{Model} & \textbf{Per iteration} & \textbf{16 iterations, $N{=}1$} \\
\midrule
Looped TF & 0.0256 & 0.410 \\
HRM & 0.0138 & 0.220 \\
TRM & 0.0256 & 0.410 \\
\textbf{GRAM (Ours)} & \textbf{0.0265} & \textbf{0.424} \\
\bottomrule
\end{tabular}
\end{table}

\paragraph{At equal compute, width is more accurate and far faster than depth.}
\cref{tab:flops_matched} compares two configurations with nearly identical FLOPs: TRM run for 320 iterations, and GRAM run for 16 iterations with $N{=}20$ samples. Because the samples are independent, they execute in a single batched pass, whereas additional recursion iterations must run one after another. Spending compute on width therefore costs far less latency than spending it on depth, and it also yields higher accuracy.

\begin{table}[h]
\caption{Matched-FLOPs comparison on Sudoku-Extreme. Latency is measured per puzzle with a single request in flight.}
\label{tab:flops_matched}
\centering
\small
\begin{tabular}{@{}lccc@{}}
\toprule
\textbf{Configuration} & \textbf{TFLOPs / puzzle} & \textbf{Latency (ms)} & \textbf{Accuracy} \\
\midrule
TRM, 320 iterations, $N{=}1$ & 8.19 & 2{,}755 & 92.15 \\
\textbf{GRAM, 16 iterations, $N{=}20$} & 8.48 & \textbf{184} & \textbf{96.06} \\
\midrule
Relative & $+3.5\%$ & $\mathbf{15.0\times}$ \textbf{faster} & $\mathbf{+3.9}$ \textbf{pp} \\
\bottomrule
\end{tabular}
\end{table}

\paragraph{GRAM reaches a given accuracy with less compute.}
\cref{tab:flops_target} reports the compute at which each model first reaches a target accuracy, obtained by sweeping the number of recursion iterations and interpolating log-linearly between measured points; we report targets that both models reach within the measured range.

\begin{table}[h]
\caption{Inference compute in TFLOPs per puzzle required to reach a target accuracy on Sudoku-Extreme.}
\label{tab:flops_target}
\centering
\small
\begin{tabular}{@{}lccc@{}}
\toprule
\textbf{Target accuracy} & \textbf{TRM} & \textbf{GRAM ($N{=}1$)} & \textbf{Savings} \\
\midrule
85 & 0.566 & 0.357 & $1.6\times$ \\
88 & 1.368 & 0.536 & $2.6\times$ \\
90 & 2.802 & 0.750 & $\mathbf{3.7\times}$ \\
\bottomrule
\end{tabular}
\end{table}

\paragraph{Numerical values behind \cref{fig:scaling}.}
\cref{tab:scaling_acc} reports the accuracy for every configuration shown in \cref{fig:scaling}. The corresponding inference compute is the product of the number of iterations, the per-iteration cost in \cref{tab:flops_per_iter}, and the number of samples $N$.

\begin{table}[h]
\caption{Accuracy (\%) on Sudoku-Extreme as a function of recursion iterations, with the corresponding inference cost. Mean $\pm$ standard deviation over 3 seeds. The compute for any entry is the number of iterations times the per-iteration cost, which for GRAM includes all $N$ samples.}
\label{tab:scaling_acc}
\centering
\small
\setlength{\tabcolsep}{5pt}
\begin{tabular}{@{}lcccccc@{}}
\toprule
\textbf{Model} & \textbf{TFLOPs / iter} & \textbf{8} & \textbf{16} & \textbf{32} & \textbf{128} & \textbf{320} \\
\midrule
Looped TF & 0.0256 & 50.0\cspstd{2.5} & 61.7\cspstd{0.9} & 64.5\cspstd{0.9} & 65.8\cspstd{1.0} & 66.1\cspstd{1.0} \\
HRM & 0.0138 & 50.4\cspstd{1.1} & 56.8\cspstd{1.2} & 60.2\cspstd{1.3} & 63.8\cspstd{1.5} & 65.4\cspstd{1.6} \\
TRM & 0.0256 & 79.3\cspstd{1.2} & 83.6\cspstd{1.3} & 86.6\cspstd{1.4} & 90.4\cspstd{1.5} & 92.1\cspstd{1.4} \\
\midrule
\textbf{GRAM}, $N{=}1$ & 0.0265 & 80.2\cspstd{2.4} & 86.6\cspstd{2.2} & 90.7\cspstd{1.8} & -- & -- \\
\textbf{GRAM}, $N{=}5$ & 0.1325 & 89.8\cspstd{2.0} & 93.6\cspstd{1.4} & 95.3\cspstd{1.1} & -- & -- \\
\textbf{GRAM}, $N{=}10$ & 0.2650 & 92.4\cspstd{1.6} & 95.1\cspstd{1.1} & 96.3\cspstd{0.8} & -- & -- \\
\textbf{GRAM}, $N{=}20$ & 0.5300 & 94.1\cspstd{1.3} & 96.1\cspstd{0.9} & 96.9\cspstd{0.7} & -- & -- \\
\textbf{GRAM}, $N{=}50$ & 1.3250 & 95.4\cspstd{1.0} & 96.8\cspstd{0.7} & 97.4\cspstd{0.6} & -- & -- \\
\bottomrule
\end{tabular}
\end{table}

\subsection{Scaling with Parallel Sampling on ARC-AGI Challenge}
\label{app:scaling}
To investigate the effect of GRAM’s sampling on the ARC-AGI-1 benchmark, we measured performance without test-time augmentation. Typically, TRM achieves its reported accuracy by generating 1,000 augmentations for a single problem and performing majority voting over the results. Because this augmentation process itself creates a wide variety of samples, we isolated the specific effect of generative sampling by performing inference solely on the original problem instance and conducting majority voting over multiple sampled paths. For a fair comparison, we reproduced TRM and evaluated it under the same setting as GRAM, on the original problem instances without test-time augmentation.

\begin{figure}[ht]
\centering
\includegraphics[width=0.7\linewidth]{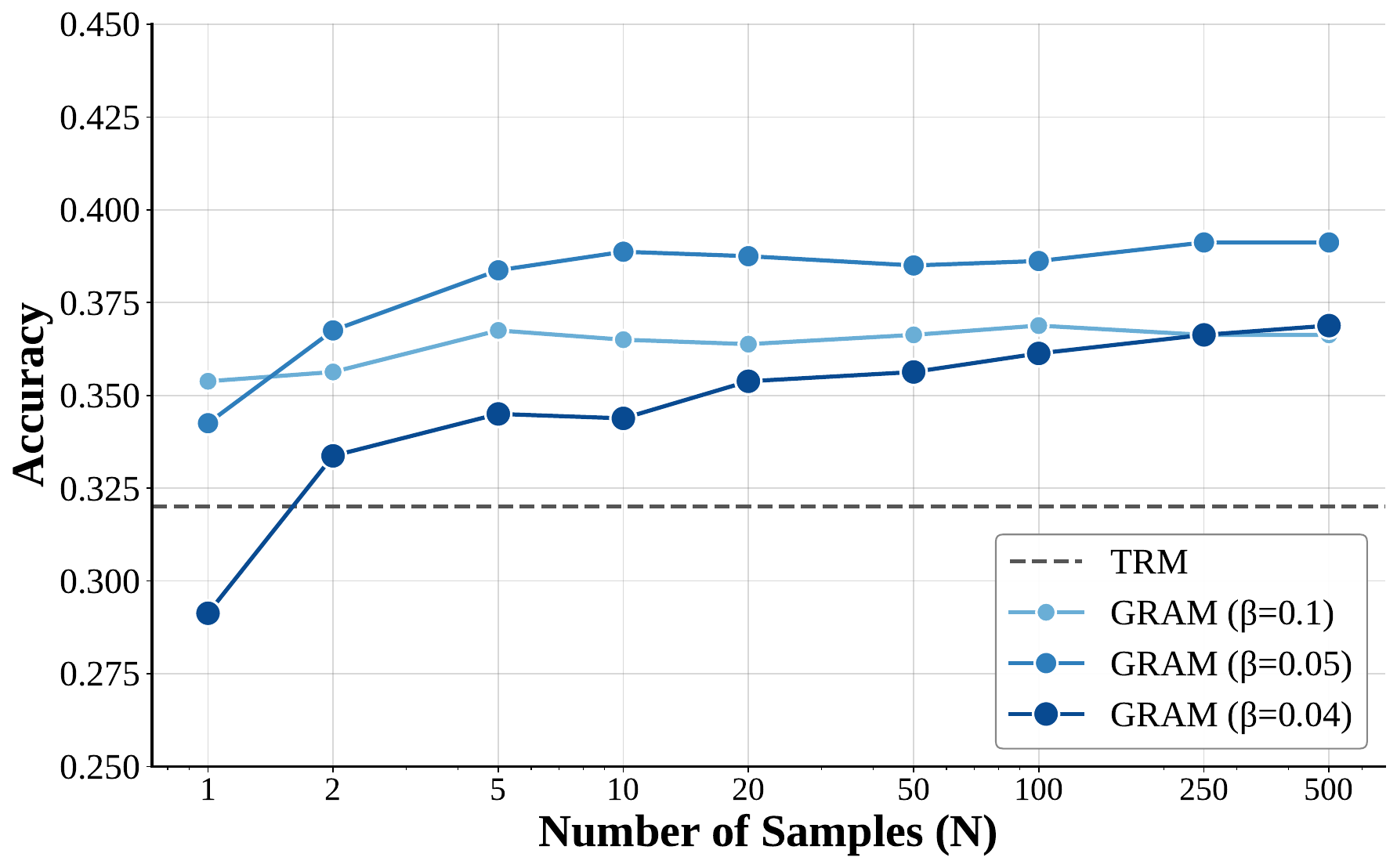}
  \caption{\textbf{Effect of sampling on ARC-AGI-1 without test-time augmentation.} To isolate the internal sampling effect, GRAM and TRM are evaluated on original problem instances without 1,000 augmentations. GRAM is shown at three KL coefficients $\beta$. While removing augmentations causes an initial performance drop, GRAM scales consistently with the number of parallel samples $N$ at every $\beta$.}
\label{fig:scaling_arc1}
\end{figure}

As illustrated in Figure~\ref{fig:scaling_arc1}, without test-time augmentation, both GRAM and TRM fall below the values reported in Table~\ref{tab:puzzle}. However, in the case of GRAM, we observe that accuracy consistently improves as the model generates more parallel samples. This trend mirrors observations in Section~\ref{sec:puzzle}, suggesting that increased inference-time compute through width scaling allows the model to explore more plausible reasoning trajectories and recover from initial errors, eventually leading to more robust solution discovery.

\begin{figure}[ht]
\centering
\includegraphics[width=0.8\linewidth]{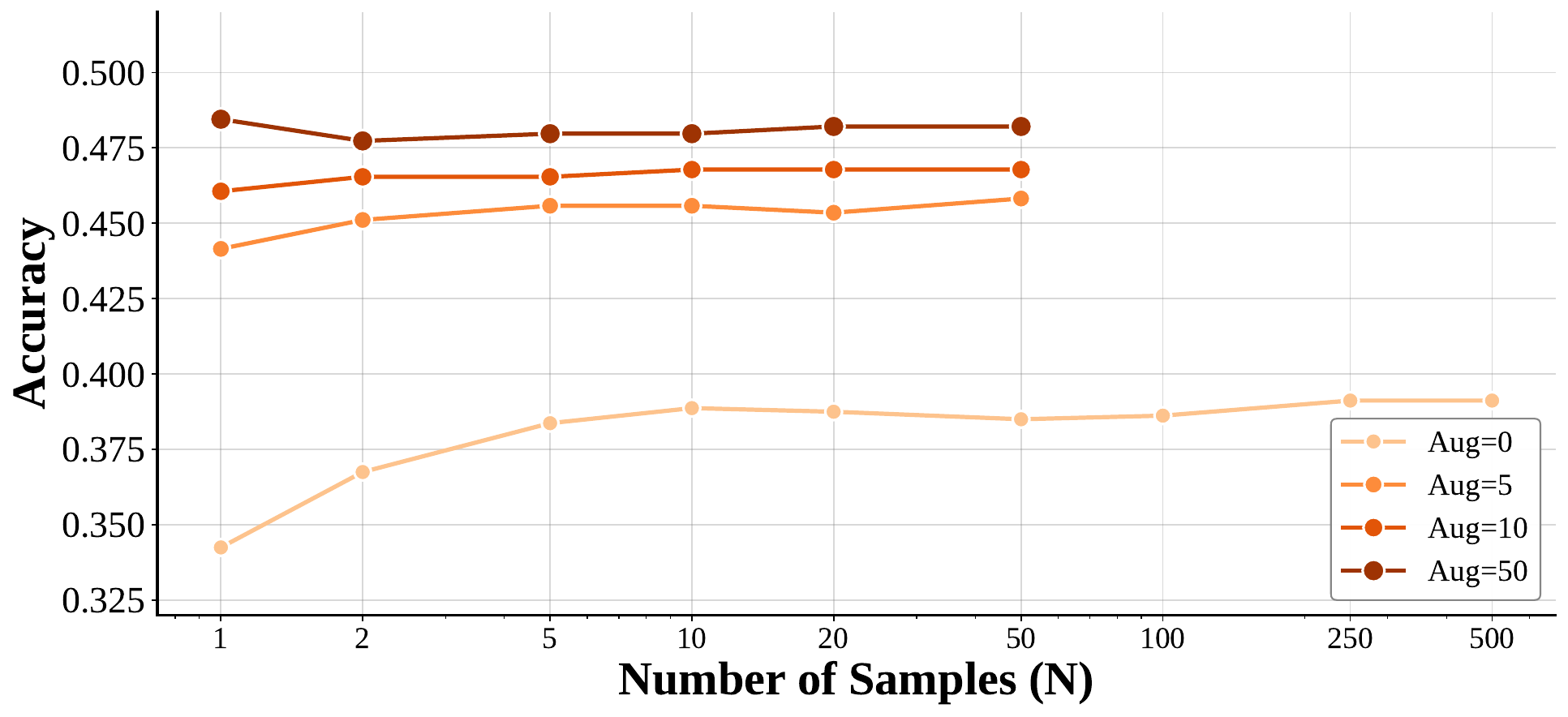}
  \caption{\textbf{Effect of augmentation on sampling efficiency.} With limited augmentation (Aug=0), parallel sampling provides consistent gains. As augmentation increases, sampling benefits diminish—at Aug$=50$, performance saturates regardless of sample count, suggesting that both provide multiple predictions for voting, and sampling does so without task-specific transformations.}
\label{fig:aug_saturation}
\end{figure}

\paragraph{Interaction between Augmentation and Sampling.}
A natural question arises: why not combine higher levels of augmentation with extensive parallel sampling? To address this, we conducted an ablation study examining the interaction between test-time augmentation and parallel sampling. Figure~\ref{fig:aug_saturation} presents the results for GRAM with $\beta=0.05$ across varying augmentation levels (Aug=0 to Aug=50).
Without augmentation (Aug=0), increasing the number of samples yields consistent accuracy improvements, demonstrating that stochastic sampling effectively explores diverse reasoning trajectories. However, as the level of augmentation increases, the marginal benefit of additional sampling diminishes substantially. At Aug$=$50, performance saturates regardless of sample count—accuracy remains nearly constant whether we draw 1 or 50 samples.
This observation reveals that augmentation and sampling are not additive, because both produce multiple predictions per input and aggregate them by voting. Augmentation obtains these predictions from hand-designed, task-specific transformations of the input, while sampling obtains them from the stochastic transitions of a single model without such design. When test-time augmentation is limited, parallel sampling compensates by exploring varied reasoning paths at inference time. When test-time augmentation is abundant, the aggregated predictions over augmented inputs already provide sufficient diversity, rendering additional inference-time exploration redundant. Consequently, scaling sampling beyond augmentation provides diminishing returns, justifying our experimental design choice to evaluate these two scaling axes separately.

\subsection{Analysis of Model Performance by Number of Valid Solutions}
\label{appx:coverage}
We group N-Queens test instances by the number of valid solutions each input admits and report single-sample accuracy and solution coverage for each group.

\begin{figure}[ht]
    \centering
    \includegraphics[width=0.5\textwidth]{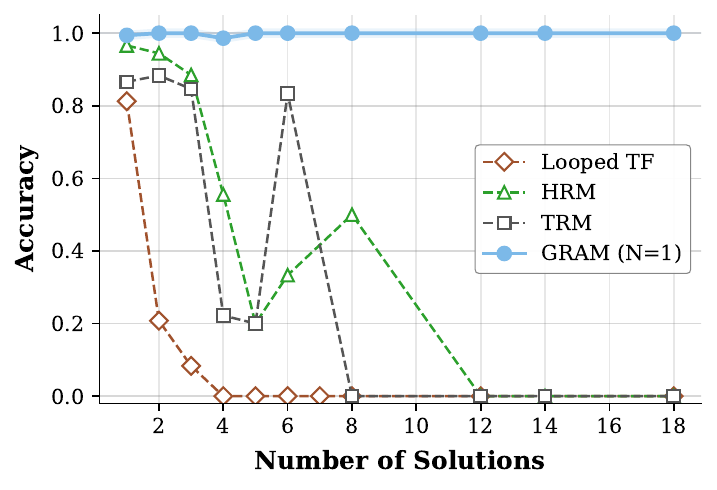}
    \caption{\textbf{Accuracy across the number of valid solutions in N-Queens ($\mathbf{8\times8}$).} Conventional deterministic recursive models suffer a sharp performance drop as the number of possible solutions increases, whereas GRAM maintains consistent performance with a single sample ($N=1$).}
    \label{fig:nqueens_accuracy}
\end{figure}
\textbf{Accuracy.} Figure~\ref{fig:nqueens_accuracy} shows single-sample accuracy on $8\times8$ N-Queens. All three deterministic recursive baselines degrade steadily as the number of valid solutions per input grows, whereas GRAM remains flat across the entire range. Since this comparison uses a single trajectory for GRAM, the gap does not come from parallel sampling. It instead reflects the training signal: inputs with more valid solutions carry more conflicting targets, which a deterministic model must collapse into one output, whereas GRAM absorbs the ambiguity in its latent variable.

\begin{figure}[ht]
    \centering
    % 첫 번째 이미지 (왼쪽)
    \begin{subfigure}[b]{0.49\textwidth}
        \centering
        \includegraphics[width=\linewidth]{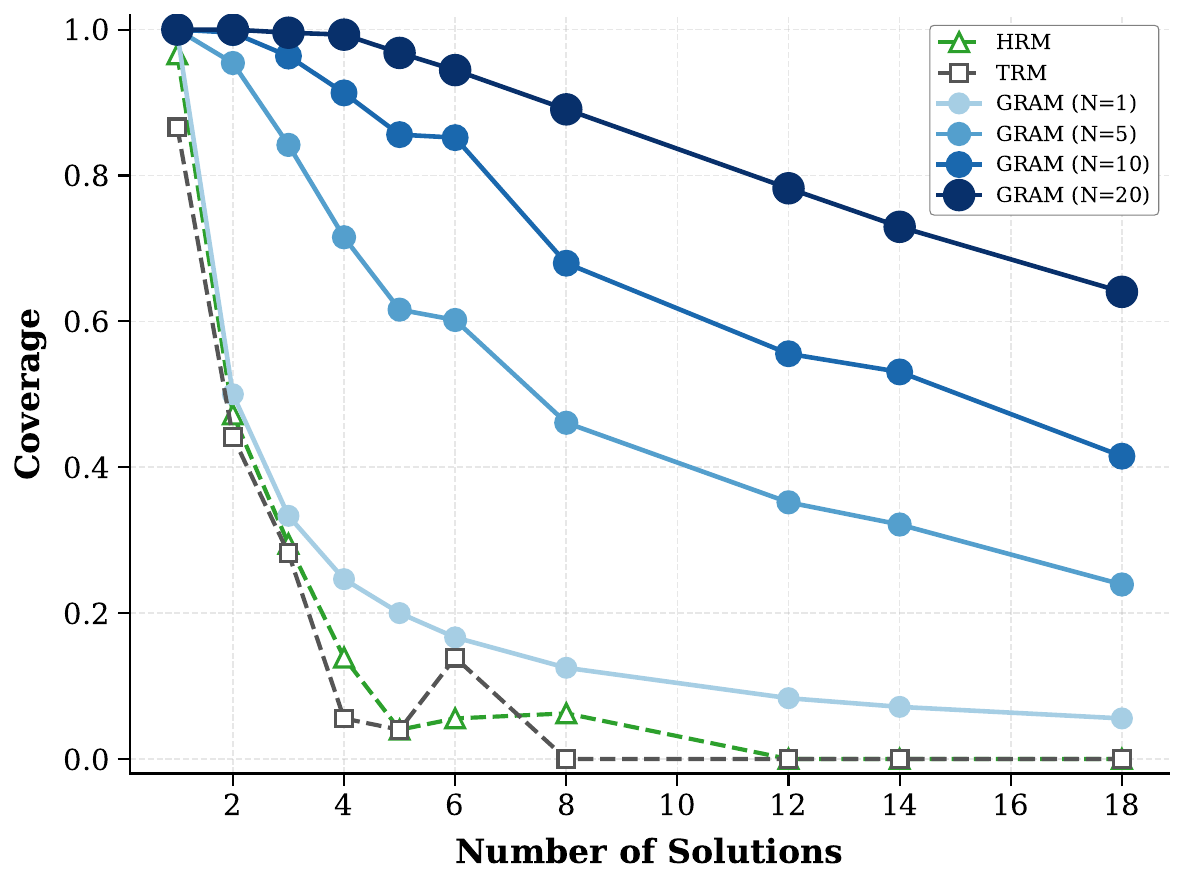}
        \caption{N-Queens $8\times8$}
        \label{fig:nqueens_8x8_coverage}
    \end{subfigure}
    \hfill % 두 이미지 사이의 간격을 자동으로 조절 (양쪽 정렬)
    % 두 번째 이미지 (오른쪽)
    \begin{subfigure}[b]{0.49\textwidth}
        \centering
        \includegraphics[width=\linewidth]{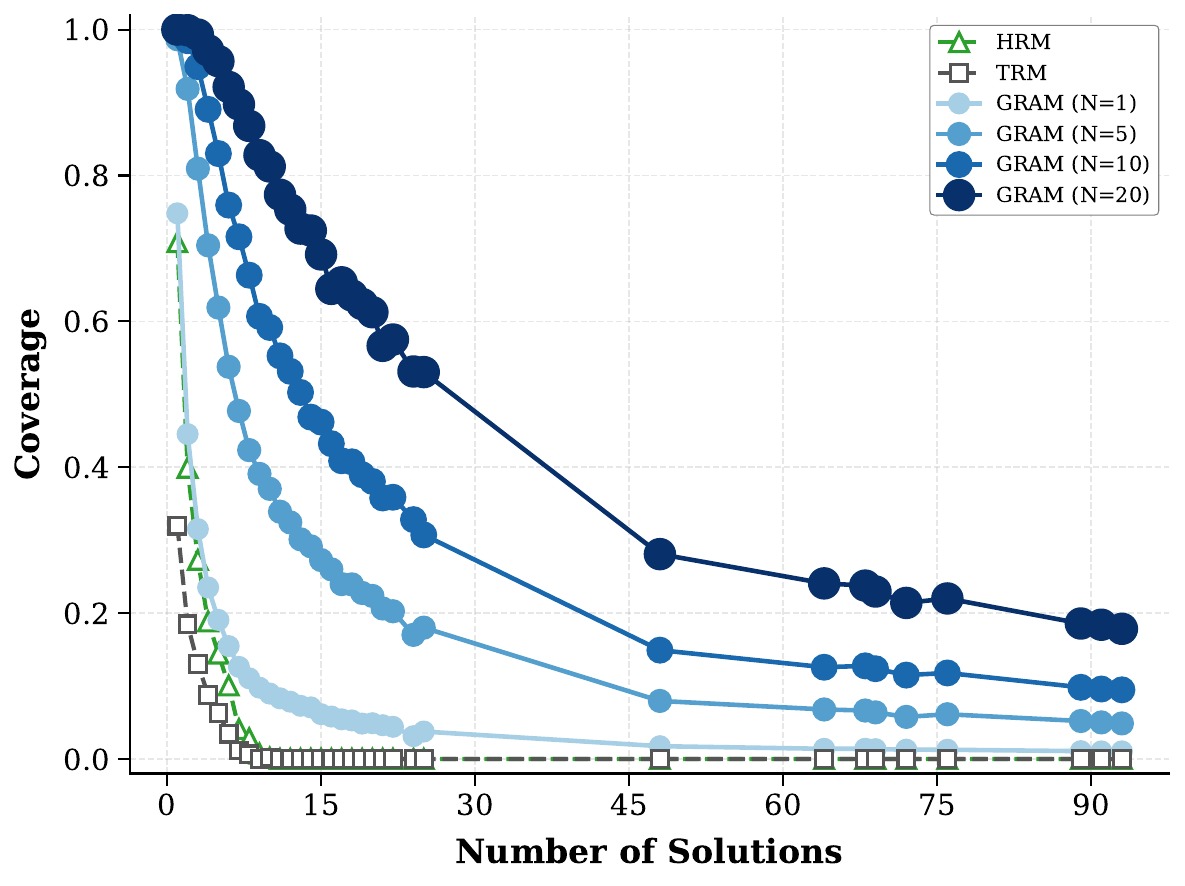}
        \caption{N-Queens $10\times10$}
        \label{fig:nqueens_10x10_coverage}
    \end{subfigure}
    
    \caption{\textbf{Solution coverage analysis on N-Queens ($\mathbf{8\times8}$ and $\mathbf{10\times10}$)} with respect to the number of ground-truth solutions. While deterministic baselines (HRM, TRM) suffer from mode collapse as the solution space grows, GRAM demonstrates monotonic improvement in coverage as the number of parallel samples $N$ increases.} 
    \label{fig:nqueens_comparison}
\end{figure}

\textbf{Coverage.} Figure~\ref{fig:nqueens_comparison} shows solution coverage on $8\times8$ and $10\times10$ N-Queens. Deterministic recursive models (HRM and TRM) decline sharply as the number of valid solutions increases, since a single fixed reasoning trajectory cannot reach alternative solutions no matter how long it is refined. GRAM instead improves monotonically as the number of parallel samples $N$ grows from 1 to 20, in both settings. Recovering multiple solutions therefore requires sampling from a distribution over trajectories rather than refining a single one, with \cref{tab:ablation_mech} isolating the contribution of stochastic guidance itself.

\subsection{Additional Generated Samples}
\label{app:generation}

\begin{figure}[ht]
\centering
\includegraphics[width=0.9\linewidth]{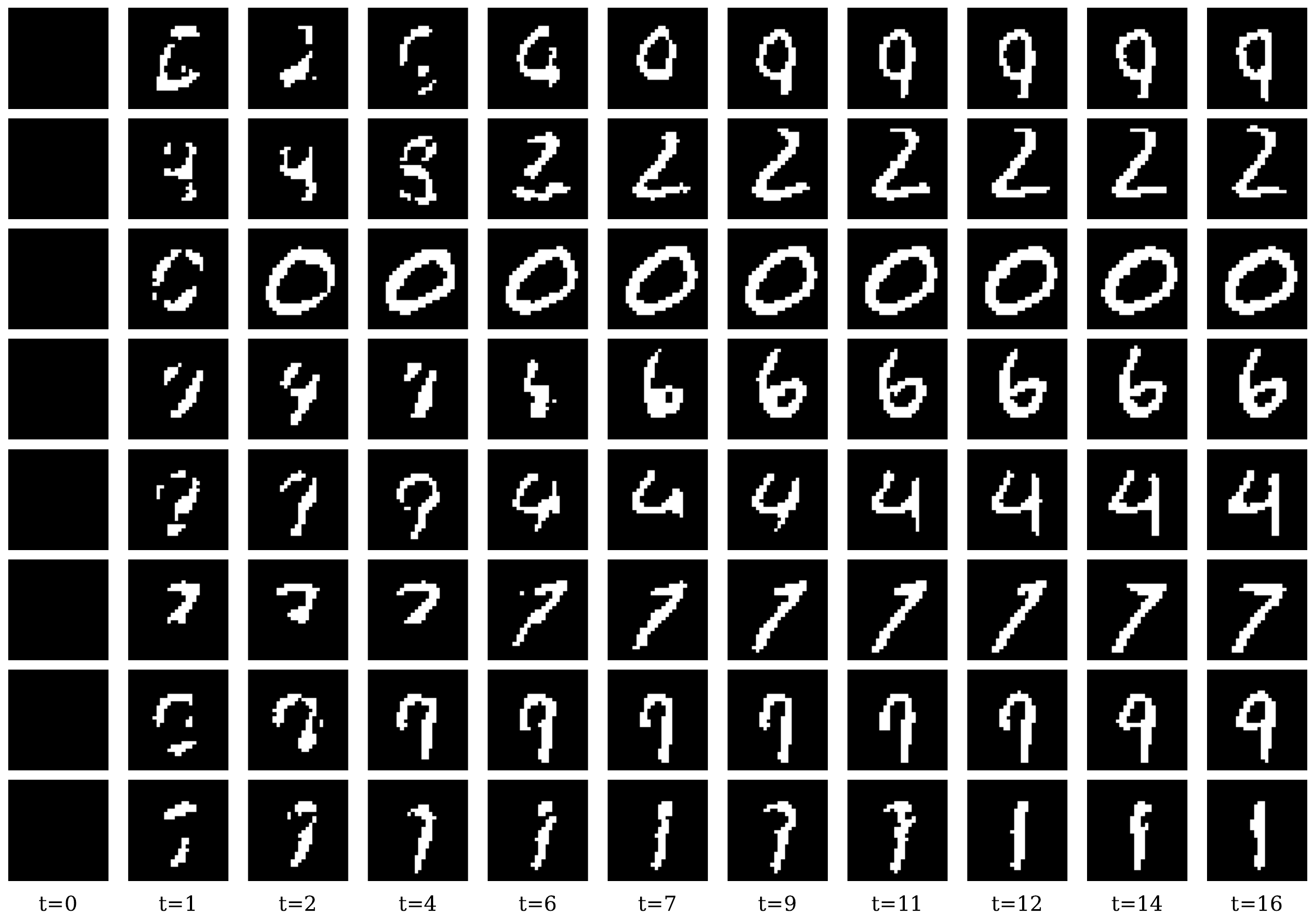}
\caption{\textbf{Additional generated samples from GRAM.} We provide 8 additional samples generated unconditionally on binarized MNIST using GRAM. Each row represents a single generated sample, visualized across its recursive refinement process.}
\label{fig:mnist_multi_sol}
\end{figure}

\paragraph{GRAM samples and progressive refinement.}
In this section, we provide further qualitative results demonstrating GRAM's capability in unconditional image generation. Figure~\ref{fig:mnist_multi_sol} presents a diverse set of samples generated on the binarized MNIST dataset, visualized across the recursive inference steps $t=0$ to $t=16$.

As observed in the main text, GRAM exhibits a distinct progressive refinement behavior. Starting from a black initialization, the model iteratively adds details and sharpens the structure of the digit. A particularly compelling property of this process is the model's ability to recover from initially ambiguous or incorrect formations. 

For instance, in the second row (generating the digit '2') and the last row (generating the digit '1'), the early predictions at $t=1$ and $t=2$ manifest as disjointed artifacts or incorrect shapes. However, as the recursion proceeds, GRAM effectively leverages its feedback loop to correct these initial errors, resolving the ambiguity and converging to a coherent, high-quality digit by $t=16$. 

\begin{figure}[ht]
\centering
\includegraphics[width=0.9\linewidth]{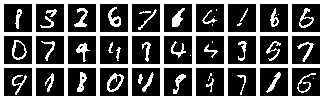}
\caption{\textbf{Unconditional samples from the VAE baseline.} Samples generated by the VAE reported in \cref{tab:generation}.}
\label{fig:vae_vis}
\end{figure}

\paragraph{VAE baseline samples.}
\label{app:vae_baseline}
For reference, \cref{fig:vae_vis} shows unconditional samples from the VAE baseline reported in \cref{tab:generation}. These samples are generated by the BCE loss variant used in the main comparison to maintain a consistent discrete-pixel representation across models.

\subsection{Additional Results on Unconditional Sudoku Generation}
\label{subsec:sudoku_generation}

As shown in Table~\ref{tab:sudoku_generation}, GRAM achieves 99.05\% validity, outperforming all D3PM baselines. The D3PM (Big) baseline reaches 79.18\% validity while using 55.1M parameters and 1000 denoising steps. In contrast, GRAM uses fewer parameters and only 16 inference steps. In all cases, the valid samples are unique under exact board matching, indicating that the reported validity is not due to simple repetition of a small set of boards. These results show that GRAM can generate highly constrained symbolic structures from an empty input, supporting its potential as a generator beyond conditional puzzle solving.

\begin{table}[h]
\caption{\textbf{Unconditional Sudoku generation.}
We report the ratio of generated boards satisfying Sudoku constraints over 100K samples. All valid boards are unique for all methods in this evaluation.}
\label{tab:sudoku_generation}
\vspace{2mm}
\centering
\small
\begin{tabular}{@{}lccc@{}}
\toprule
\textbf{Method} & \textbf{\#Params} & \textbf{Steps} & \textbf{Validity(\%)} \\
\midrule
D3PM (Big) & 55.1M & 1000 & 79.18 \\
D3PM (Small) & 15.9M & 1000 & 21.88 \\
\rowcolor{blue!8}
\textbf{GRAM (Ours)} & \textbf{10.9M} & \textbf{16} & \textbf{99.05} \\
\bottomrule
\end{tabular}
\end{table}

\subsection{Stochasticity and Guidance Ablation without Parallel Sampling}
\label{app:ablation_n1}

Table~\ref{tab:ablation_mech} in the main text reports the stochasticity and guidance ablation at $N$=20. Table~\ref{tab:ablation_mech_full} adds the single-sample results, so that the contribution of stochasticity and learned guidance can be read separately from the effect of parallel sampling. TRM with stochastic decoding samples only the output tokens, and on Sudoku its output distribution is nearly one-hot, so its samples match the argmax prediction and its two Sudoku columns coincide.

\begin{table}[h]
    \caption{\textbf{Stochasticity and guidance ablation at $N$=1 and $N$=20} on Sudoku-Extreme and N-Queens ($8\times8$). Values are means over 3 seeds with standard deviations.}
    \label{tab:ablation_mech_full}
    \centering
    \footnotesize
    \setlength{\tabcolsep}{6pt}
    \renewcommand{\arraystretch}{1.1}
    \begin{tabular}{@{}lcccc@{}}
    \toprule
    & \multicolumn{2}{c}{\textbf{Sudoku}} & \multicolumn{2}{c}{\textbf{N-Queens}} \\
    \cmidrule(lr){2-3} \cmidrule(lr){4-5}
    \textbf{Model variant} & $N$=1 & $N$=20 & $N$=1 & $N$=20 \\
    \midrule
    \textbf{GRAM (ours)} & 86.6$_{\pm2.2}$ & \textbf{96.1}$_{\pm0.9}$ & 99.5$_{\pm0.3}$ & \textbf{100.0}$_{\pm0.0}$ \\
    \midrule
    w/o stochastic guidance & 83.6$_{\pm1.3}$ & 83.6$_{\pm1.3}$ & 66.8$_{\pm5.6}$ & 66.8$_{\pm5.6}$ \\
    stochasticity only & 86.6$_{\pm1.4}$ & 96.0$_{\pm1.1}$ & 60.2$_{\pm15.1}$ & 62.1$_{\pm14.9}$ \\
    guide only & 0.0$_{\pm0.0}$ & 0.0$_{\pm0.0}$ & 0.0$_{\pm0.0}$ & 0.0$_{\pm0.0}$ \\
    w/ direct prediction & 44.1$_{\pm5.2}$ & 70.7$_{\pm7.6}$ & 97.5$_{\pm3.5}$ & 98.8$_{\pm2.2}$ \\
    \midrule
    TRM w/ stochastic decoder & 79.7$_{\pm1.9}$ & 79.7$_{\pm1.9}$ & 56.8$_{\pm5.7}$ & 65.9$_{\pm4.7}$ \\
    TRM w/ random init. & 79.9$_{\pm1.2}$ & 91.2$_{\pm0.9}$ & 71.1$_{\pm8.5}$ & 71.4$_{\pm8.5}$ \\
    \bottomrule
    \end{tabular}
\end{table}

\subsection{Placement of the Stochastic Transition}
\label{app:noise-placement}

The hierarchical instantiation admits a stochastic transition at either level, the high-level update or the low-level refinement. We compare three low-level variants with our high-level design on N-Queens $8\times8$. \emph{Every-low} samples $\epsilon$ at all $K$ low-level refinements, trained either with the KL of the final refinement only (Last KL) or with the KL averaged over the $K$ refinements (Mean KL). \emph{Last-low} samples $\epsilon$ once, at the final low-level refinement, and uses only that KL term.

\begin{table}[h]
\small
\centering
\caption{Placement of the stochastic transition in GRAM on N-Queens $8\times8$.
Accuracy is single-sample and coverage uses 20 samples.}
\label{tab:noise-placement}
\begin{tabular}{lcc}
\toprule
\textbf{Placement} & \textbf{Accuracy} & \textbf{Coverage} \\
\midrule
High-level (default) & $99.5 \pm 0.3$ & $89.7 \pm 1.7$ \\
Every-low, Last KL & $0.0 \pm 0.0$  & $0.0 \pm 0.0$ \\
Every-low, Mean KL & $0.0 \pm 0.0$  & $0.0 \pm 0.0$ \\
Last-low, Last KL  & $99.2 \pm 0.6$ & $90.2 \pm 0.4$ \\
\bottomrule
\end{tabular}
\end{table}

\cref{tab:noise-placement} shows that Every-low fails under both KL variants, while Last-low matches our high-level design. This suggests that the number of injections per transition matters more than the level at which they occur: Every-low draws from the posterior $K$ times within a single high-level update ($K{=}4$ here), so each transition receives $K$ times as much target information, which we attribute to the resulting overfitting. Our formulation defines the stochastic transition on $z_t$ and is agnostic to placement, so we keep the injection at the high level, which trains stably.

\subsection{Reinforcement Learning Fine-Tuning of GRAM}
\label{app:rl_finetuning}

We investigate whether GRAM can be further improved with outcome-level feedback by updating its stochastic transition prior using proximal policy optimization (PPO)~\citep{schulman2017proximal}, while freezing the remaining pretrained parameters.

\textbf{Setup.}
At each recursive step, we treat the stochastic residual
$\epsilon_t \sim p_\theta(\epsilon_t \mid u_t)$ as the policy action that determines the next latent transition. After each transition, we assign a binary reward of one if the current prediction is a valid N-Queens solution and zero otherwise. A trajectory terminates upon finding a valid solution or after 16 steps. We train for 1,000 PPO updates with a batch of 512 trajectories, using eight trajectories per problem. We use $\gamma=0.99$, GAE~\citep{schulman2015high} $\lambda=0.95$, a clipping ratio of 0.2, two PPO epochs, and learning rates of $3\times10^{-5}$ and $10^{-4}$ for the transition prior and PPO critic, respectively.

\textbf{Results.}
\cref{tab:nqueens_rl} reports single-sample accuracy ($N{=}1$), majority-vote accuracy with 20 samples ($N{=}20$), and coverage using the same 20 samples. On $8\times8$ N-Queens, GRAM is already near saturation, and reinforcement learning provides modest additional gains. On $10\times10$ N-Queens, GRAM-RL improves single-sample accuracy from 89.7\% to 99.1\% and coverage from 57.5\% to 61.8\%, while majority-vote accuracy remains saturated.

\begin{table}[h]
    \caption{\textbf{Reinforcement learning fine-tuning on N-Queens.}
    We report single-sample accuracy ($N{=}1$), majority-vote accuracy with 20 samples ($N{=}20$), and coverage with 20 samples. Values are means $\pm$ sample standard deviations over 3 seeds.}
    \label{tab:nqueens_rl}
    \centering
    \small
    \providecommand{\cspstd}[1]{{\scriptsize$\pm#1$}}
    \adjustbox{max width=\textwidth}{
    \setlength{\tabcolsep}{6pt}
    \begin{tabular}{@{}lcccccc@{}}
        \toprule
        & \multicolumn{3}{c}{\textbf{N-Queens} ($8\times8$)}
        & \multicolumn{3}{c}{\textbf{N-Queens} ($10\times10$)} \\
        \cmidrule(lr){2-4} \cmidrule(lr){5-7}
        \textbf{Method}
        & $N{=}1$ & $N{=}20$ & Coverage
        & $N{=}1$ & $N{=}20$ & Coverage \\
        \midrule
        GRAM (Ours)
        & 99.5\cspstd{0.3}
        & 100.0\cspstd{0.0}
        & 89.7\cspstd{1.7}
        & 89.7\cspstd{2.7}
        & 100.0\cspstd{0.0}
        & 57.5\cspstd{3.4} \\
        \rowcolor{blue!8}
        \textbf{GRAM-RL}
        & \textbf{99.9}\cspstd{0.2}
        & 100.0\cspstd{0.0}
        & \textbf{90.7}\cspstd{0.5}
        & \textbf{99.1}\cspstd{0.3}
        & 100.0\cspstd{0.0}
        & \textbf{61.8}\cspstd{0.2} \\
        \bottomrule
    \end{tabular}
    }
\end{table}

\subsection{Visualizing Latent Recursion Process}
\label{sec:trajectory}
To understand how stochastic guidance shapes reasoning, we visualize latent trajectories during recursive computation. Specifically, we track the high-level state $h$ at each supervision step throughout the recursion process. For visualization, we project these latent vectors into 2D using PCA~\citep{shlens2014tutorial} and interpolate unobserved states via K-D tree~\citep{kdtree} to construct a continuous loss landscape.

\begin{figure}[ht]
\centering
\includegraphics[width=0.9\linewidth]{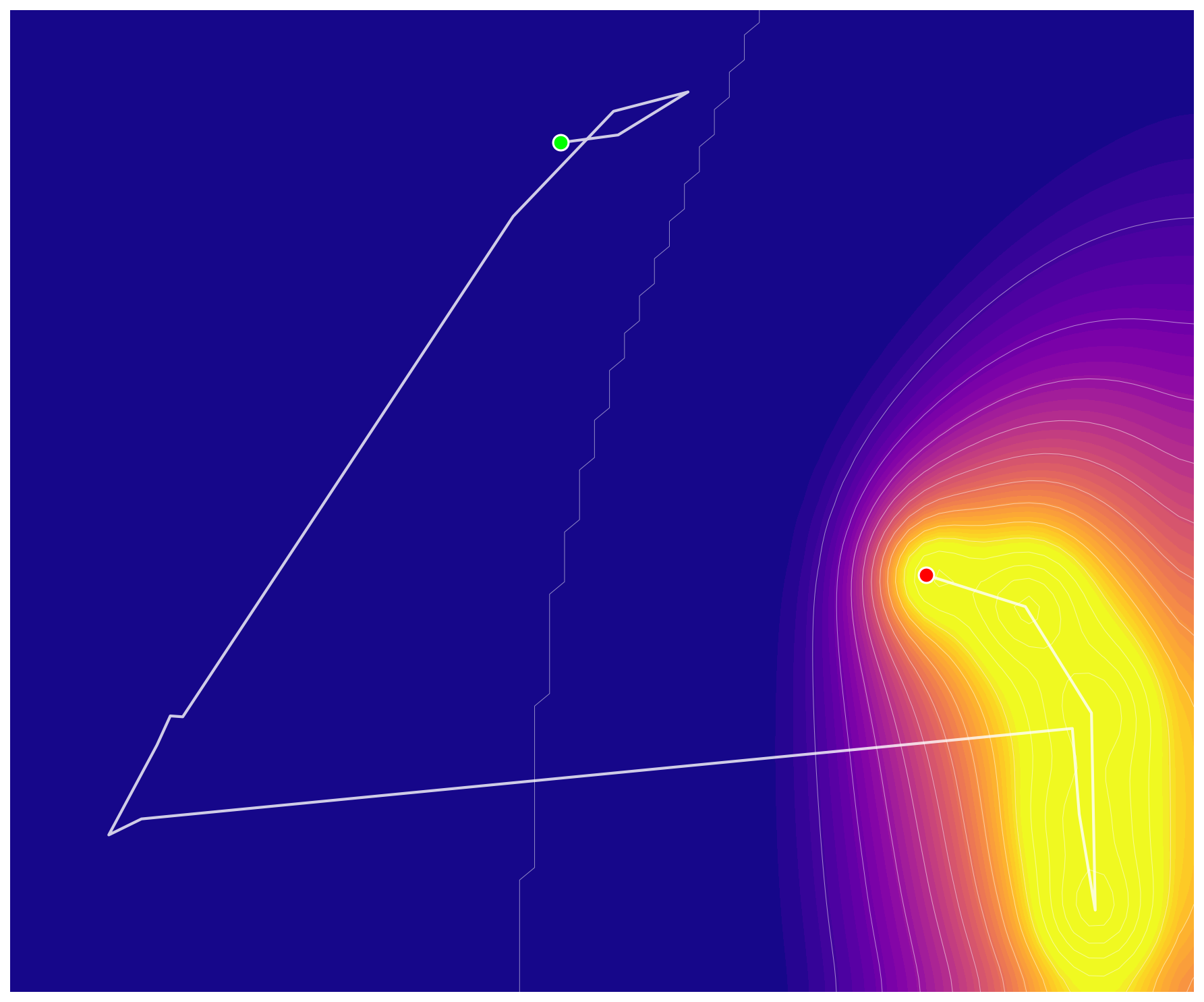}
\caption{\textbf{Latent reasoning trajectory of TRM.} The red dot indicates the initial state $h_0$ and the green dot indicates the final state $h_T$. Background color represents the loss landscape: bright yellow corresponds to high loss regions, while dark blue indicates low loss (optimal) regions. TRM follows a single deterministic path with no ability to escape suboptimal trajectories.}
\label{fig:trm_traj}
\end{figure}

\begin{figure}[ht]
\centering
\includegraphics[width=0.9\linewidth]{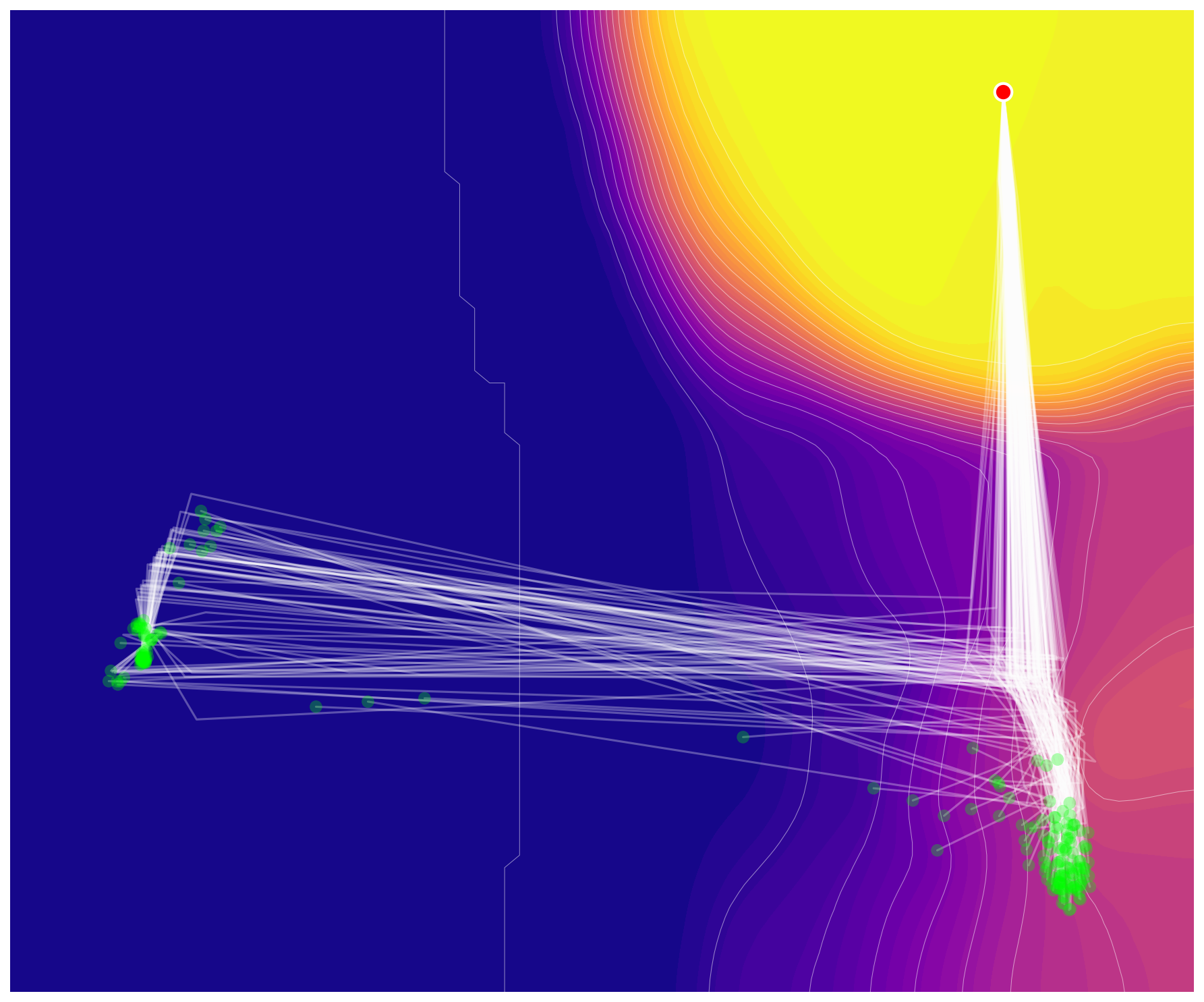}
\caption{\textbf{Latent reasoning trajectories of GRAM (50 samples).} Using the same visualization scheme as Figure~\ref{fig:trm_traj}, we show 50 sampled trajectories from GRAM. The stochastic guidance enables diverse exploration of the latent space: while some trajectories converge to local minima (right bottom), others successfully reach the global optimum (left middle), demonstrating how parallel sampling improves solution discovery.}
\label{fig:gram_traj}
\end{figure}

Figures~\ref{fig:trm_traj} and~\ref{fig:gram_traj} compare TRM and GRAM on the same Sudoku puzzle. TRM follows a single deterministic path from initialization to solution, offering no mechanism to escape if the trajectory enters a suboptimal region. In contrast, GRAM samples diverse trajectories that explore different regions of latent space before converging. While some trajectories become trapped in local minima (bottom right), others successfully navigate toward the global optimum (dark blue regions). This diversity enables GRAM to discover valid solutions more reliably through parallel exploration.

\clearpage

\section*{Licenses}

\begin{center}
\small
\captionof{table}{\textbf{Existing assets, licenses, and source links.}
We list the existing datasets, benchmarks, and public reference implementations
used or cited in our experiments. Synthetic N-Queens and Graph Coloring instances
are generated by the authors and are therefore not external assets.}
\label{tab:asset_licenses}

\begin{adjustbox}{max width=\textwidth}
\begin{tabular}{p{0.18\linewidth} p{0.23\linewidth} p{0.16\linewidth} p{0.34\linewidth}}
\toprule
\textbf{Asset} & \textbf{Use in this paper} & \textbf{License / terms} & \textbf{Source link} \\
\midrule
MNIST & Binarized MNIST generation experiments & Creative Commons Attribution-Share Alike 3.0 & \url{https://keras.io/api/datasets/mnist/} \\
ARC-AGI-1 / original ARC & ARC-AGI reasoning benchmark & Apache License 2.0 & \url{https://github.com/fchollet/ARC-AGI} \\
ARC-AGI-2 & ARC-AGI-2 reasoning benchmark / reference results & Apache License 2.0 & \url{https://github.com/arcprize/ARC-AGI-2} \\
HRM repository & HRM baseline and Sudoku-Extreme-related reference implementation & Apache License 2.0 & \url{https://github.com/sapientinc/HRM} \\
TinyRecursiveModels / TRM repository & TRM baseline and recursive reasoning reference implementation & MIT License & \url{https://github.com/SamsungSAILMontreal/TinyRecursiveModels} \\
MDLM repository & Masked diffusion baseline reference implementation & Apache License 2.0 & \url{https://github.com/kuleshov-group/mdlm} \\
Google Research D3PM implementation & D3PM image-generation baseline reference implementation & Apache License 2.0 & \url{https://github.com/google-research/google-research/blob/master/d3pm/images/diffusion_categorical.py} \\
Looped Transformer repository & Looped Transformer baseline reference implementation & MIT License & \url{https://github.com/Leiay/looped_transformer} \\
N-Queens & Synthetic multi-solution constraint satisfaction task generated by the authors & Not an external asset & N/A \\
Graph Coloring & Synthetic multi-solution constraint satisfaction task generated by the authors & Not an external asset & N/A \\
\bottomrule
\end{tabular}
\end{adjustbox}
\end{center}

\clearpage

\ifchecklist
\newpage

\input{checklist_camera.tex}
\fi

\end{document}

%% file: checklist_camera.tex
\section*{NeurIPS Paper Checklist}

\begin{enumerate}

\item {\bf Claims}
    \item[] Question: Do the main claims made in the abstract and introduction accurately reflect the paper's contributions and scope?
    \item[] Answer: \answerYes{} % Replace by \answerYes{}, \answerNo{}, or \answerNA{}.
    \item[] Justification: The abstract and introduction state the main claims of GRAM as a probabilistic recursive reasoning framework, including stochastic latent trajectories, width-based inference-time scaling, and empirical gains over deterministic recursive baselines. These claims are supported by the method and experiments in Sections~\ref{sec:architecture}--\ref{sec:experiments} and the appendix.

\item {\bf Limitations}
    \item[] Question: Does the paper discuss the limitations of the work performed by the authors?
    \item[] Answer: \answerYes{} % Replace by \answerYes{}, \answerNo{}, or \answerNA{}.
    \item[] Justification: The paper includes a Conclusions and Limitations section discussing the main limitation of sequential deep supervision and its impact on training efficiency and scaling. Additional scope limitations are reflected in the experimental setup, which focuses on structured reasoning and generation benchmarks rather than broad general-purpose deployment.

\item {\bf Theory assumptions and proofs}
    \item[] Question: For each theoretical result, does the paper provide the full set of assumptions and a complete (and correct) proof?
    \item[] Answer: \answerNA{} % Replace by \answerYes{}, \answerNo{}, or \answerNA{}.
    \item[] Justification: The paper does not present formal theoretical results in the form of theorems requiring complete proofs. It provides probabilistic modeling derivations and an analysis of the truncated surrogate objective in the method section and appendix.

    \item {\bf Experimental result reproducibility}
    \item[] Question: Does the paper fully disclose all the information needed to reproduce the main experimental results of the paper to the extent that it affects the main claims and/or conclusions of the paper (regardless of whether the code and data are provided or not)?
    \item[] Answer: \answerYes{} % Replace by \answerYes{}, \answerNo{}, or \answerNA{}.
    \item[] Justification: The paper describes the model architecture, training objective, datasets, evaluation protocols, baselines, and hyperparameters in the main text and appendix. The code is released together with the camera-ready version.

\item {\bf Open access to data and code}
    \item[] Question: Does the paper provide open access to the data and code, with sufficient instructions to faithfully reproduce the main experimental results, as described in supplemental material?
    \item[] Answer: \answerYes{} % Replace by \answerYes{}, \answerNo{}, or \answerYes{}.
    \item[] Justification: The code is released together with the camera-ready version and is linked from the project page given in the abstract.

\item {\bf Experimental setting/details}
    \item[] Question: Does the paper specify all the training and test details (e.g., data splits, hyperparameters, how they were chosen, type of optimizer) necessary to understand the results?
    \item[] Answer: \answerYes{} % Replace by \answerYes{}, \answerNo{}, or \answerNA{}.
    \item[] Justification: The paper specifies the benchmark tasks, train/test setups, model configurations, optimization choices, evaluation metrics, and hyperparameters in the experiments section and appendix. These details are intended to make the experimental results interpretable and reproducible.
\item {\bf Experiment statistical significance}
    \item[] Question: Does the paper report error bars suitably and correctly defined or other appropriate information about the statistical significance of the experiments?
    \item[] Answer: \answerYes{}
    \item[] Justification: The paper reports error bars for the main experimental results that support the paper's claims. The error bars summarize variability across repeated runs with different random seeds, and are reported as mean $\pm$ standard deviation over 3 seeds.

\item {\bf Experiments compute resources}
    \item[] Question: For each experiment, does the paper provide sufficient information on the computer resources (type of compute workers, memory, time of execution) needed to reproduce the experiments?
    \item[] Answer: \answerYes{} % Replace by \answerYes{}, \answerNo{}, or \answerNA{}.
    \item[] Justification: Compute resources are reported in Appendix~\ref{appx:train_detail}, including the GPU type, number of GPUs, and wall-clock training time for each task. Experiments were conducted using NVIDIA RTX 4090 and RTX 6000 PRO GPUs, and all reported training and inference times are measured on RTX 4090 GPUs for comparability.
    
\item {\bf Code of ethics}
    \item[] Question: Does the research conducted in the paper conform, in every respect, with the NeurIPS Code of Ethics \url{https://neurips.cc/public/EthicsGuidelines}?
    \item[] Answer: \answerYes{} % Replace by \answerYes{}, \answerNo{}, or \answerNA{}.
    \item[] Justification: The authors reviewed the NeurIPS Code of Ethics and the research conforms to it. The work uses standard public benchmarks and synthetic structured reasoning tasks, and does not involve human subjects or sensitive personal data.

\item {\bf Broader impacts}
    \item[] Question: Does the paper discuss both potential positive societal impacts and negative societal impacts of the work performed?
    \item[] Answer: \answerYes{} % Replace by \answerYes{}, \answerNo{}, or \answerNA{}.
    \item[] Justification: The paper discusses potential positive impacts of more efficient recursive reasoning and potential negative impacts such as plausible but invalid generations being mistaken for verified solutions and the computational and energy costs of multi-sample inference. The work is foundational and does not propose deployment in human-facing decision systems.
    
\item {\bf Safeguards}
    \item[] Question: Does the paper describe safeguards that have been put in place for responsible release of data or models that have a high risk for misuse (e.g., pre-trained language models, image generators, or scraped datasets)?
    \item[] Answer: \answerNA{} % Replace by \answerYes{}, \answerNo{}, or \answerNA{}.
    \item[] Justification: The paper does not release high-risk pretrained language models, scraped datasets, or human-facing systems. The experiments are limited to structured reasoning benchmarks and standard image-generation benchmarks, so additional release safeguards are not applicable.
\item {\bf Licenses for existing assets}
    \item[] Question: Are the creators or original owners of assets (e.g., code, data, models), used in the paper, properly credited and are the license and terms of use explicitly mentioned and properly respected?
    \item[] Answer: \answerYes{}
    \item[] Justification: The paper cites the original sources of the datasets, benchmarks, and baselines used in the experiments. Table~\ref{tab:asset_licenses} lists the licenses and source links for the existing assets used, including MNIST, ARC-AGI, HRM/TRM-related assets, and public reference implementations where applicable.

\item {\bf New assets}
    \item[] Question: Are new assets introduced in the paper well documented and is the documentation provided alongside the assets?
    \item[] Answer: \answerYes{} % Replace by \answerYes{}, \answerYes{}, or \answerNA{}.
    \item[] Justification: The released code is documented with usage instructions and license information and is linked from the project page. The Graph Coloring dataset is released on Hugging Face, and the N-Queens dataset is generated by the released code; both constructions are described in Appendix~\ref{sec:multi_solution_details}.

\item {\bf Crowdsourcing and research with human subjects}
    \item[] Question: For crowdsourcing experiments and research with human subjects, does the paper include the full text of instructions given to participants and screenshots, if applicable, as well as details about compensation (if any)? 
    \item[] Answer: \answerNA{} % Replace by \answerYes{}, \answerNo{}, or \answerNA{}.
    \item[] Justification: The research does not involve crowdsourcing, user studies, annotation workers, or other human-subject experiments.

\item {\bf Institutional review board (IRB) approvals or equivalent for research with human subjects}
    \item[] Question: Does the paper describe potential risks incurred by study participants, whether such risks were disclosed to the subjects, and whether Institutional Review Board (IRB) approvals (or an equivalent approval/review based on the requirements of your country or institution) were obtained?
    \item[] Answer: \answerNA{} % Replace by \answerYes{}, \answerNo{}, or \answerNA{}.
    \item[] Justification: The research does not involve crowdsourcing or human-subject experiments, so IRB approval or equivalent review is not applicable.

\item {\bf Declaration of LLM usage}
    \item[] Question: Does the paper describe the usage of LLMs if it is an important, original, or non-standard component of the core methods in this research? Note that if the LLM is used only for writing, editing, or formatting purposes and does \emph{not} impact the core methodology, scientific rigor, or originality of the research, declaration is not required.
    %this research? 
    \item[] Answer: \answerNA{} % Replace by \answerYes{}, \answerNo{}, or \answerNA{}.
    \item[] Justification: Large language models were used only for writing, editing, and formatting assistance, and were not used as an important, original, or non-standard component of the core methodology, experiments, data analysis, or scientific conclusions. Under the checklist guidance, a declaration is therefore not required.

\end{enumerate}